\documentclass{article}
\usepackage{iclr2027_conference,times}

\usepackage{amsmath,amsfonts,bm}

\def\eqref#1{equation~\ref{#1}}

\def\1{\bm{1}}

\DeclareMathAlphabet{\mathsfit}{\encodingdefault}{\sfdefault}{m}{sl}
\SetMathAlphabet{\mathsfit}{bold}{\encodingdefault}{\sfdefault}{bx}{n}

\usepackage[pagebackref=false,breaklinks=true,colorlinks,bookmarks=false,citecolor=citecolor,linkcolor=linkcolor]{hyperref}
\usepackage{url}
\usepackage{graphicx}
\usepackage{booktabs}
\usepackage{amsmath,amssymb}
\usepackage{colortbl}
\definecolor{mygray}{gray}{.9}
\usepackage{xcolor}
\definecolor{mygray}{gray}{.9}
\definecolor{tablehighlight}{HTML}{F2F5F8}
\definecolor{ourshighlight}{HTML}{E8FFFF}
\usepackage{multirow}
\usepackage{enumitem}
\usepackage{xspace}
\usepackage{array}
\usepackage{subcaption}

\definecolor{citecolor}{HTML}{0071BC}
\definecolor{linkcolor}{HTML}{ED1C24}

\renewcommand{\cite}{\citep}

\newcommand{\ours}{RASteer\xspace}

\newcommand{\up}{\ensuremath{\uparrow}}
\newcommand{\dn}{\ensuremath{\downarrow}}

\newcommand{\Prr}{\mathbf{P}_r}

\graphicspath{{figures/}}

\title{\ours: Retain-Aware Activation Steering for Concept Erasure in Diffusion Models}

\author{Yongliang Wu$^{1}$, Haori Lu$^{1}$, Yulun Wu$^{1}$, Jinqi Luo$^{2}$, Xingyu Zhu$^{3}$\thanks{Corresponding author}, Yaoyao Liu$^{1}$ \\$^{1}$ University of Illinois Urbana-Champaign \\ $^{2}$ University of Pennsylvania \\ $^{3}$ National University of Singapore \\}
\iclrfinalcopy 
\begin{document}

\maketitle

\begin{abstract}
Concept erasure aims to remove a target concept, such as a copyrighted style, a recognizable character,
or unsafe content, from a pretrained text-to-image diffusion model while preserving its ability to generate
other content. Existing activation steering methods build an erasure direction mainly from the target
concept and adjust model activations along it at inference time. However, target and retained concepts
often overlap in the model's representation space, so this direction also contains shared components that
retained concepts rely on. Steering directly along this direction can therefore suppress retained concepts
and harm the generation of non-target content. To address this issue, we propose \textbf{R}etain-aware \textbf{A}ctivation
\textbf{Steer}ing (\textbf{\ours}), a training-free method. \ours\ first builds a retain subspace from the concepts to preserve.
Retain-Orthogonal Steering (ROS) then removes components aligned with this subspace from the erasure
direction, making steering more specific to the target. Since fully removing the shared components can
weaken erasure, we further introduce Overlap-Adaptive Calibration (OAC). At each layer and denoising step,
OAC uses the overlap between the erasure direction and the retain subspace to control how much of each
shared component is removed, balancing target erasure and concept preservation. Experiments on unsafe-content,
instance, and artistic-style erasure across multiple backbones and benchmarks show that \ours\ matches or
outperforms the activation steering and weight editing baselines we evaluate, achieving a better
balance between erasure and preservation.\footnote{Project page: \url{https://rasteer.cvmlgroup.web.illinois.edu/}}
\end{abstract}

\section{Introduction}
\label{sec:intro}

Text-to-image (T2I) diffusion models~\citep{rombach2022high} are trained on web-scale data and inevitably learn content that their providers cannot include in their products, such as copyrighted artistic styles, recognizable characters, and unsafe images.
When takedown requests, changes in regulations, or safety audits require removing such content, retraining on filtered data is too costly, while prompt- or output-level filters are easily bypassed~\citep{rando2022red,zhang2026concept,sarwar2026multimodal}.
Concept erasure offers a practical alternative: it directly edits a released model to prevent it from generating a target concept~\citep{gandikota2023erasing,kumari2023ablating,gandikota2024unified,gong2024reliable,lu2024mace,li2026speed}, and has become a key step in deploying T2I models responsibly.
Such edits must erase the target while preserving all other content, including unrelated styles and characters, as well as open-domain generation.

Existing erasure methods broadly fall into two categories: weight editing and activation steering.
Weight editing fine-tunes the model or applies closed-form updates to its cross-attention matrices~\citep{gandikota2023erasing,kumari2023ablating,gandikota2024unified,gong2024reliable,lu2024mace,li2026speed,liu2025rethinking,liu2026image,hu2026unibench}, writing these changes into model parameters and increasing the cost of switching or undoing edits.
By contrast, activation steering~\citep{gaintseva2026casteer,huu2024effects,xie2025erasing,zhang2024steerdiff} keeps model weights fixed and suppresses target concepts by adjusting intermediate activations during inference, allowing edits to be enabled, disabled, or combined.
The erasure direction is constructed from the target concept, and each image patch is gated based on its activation's alignment with that direction.

\begin{figure}[t]
\centering
\includegraphics[width=\linewidth]{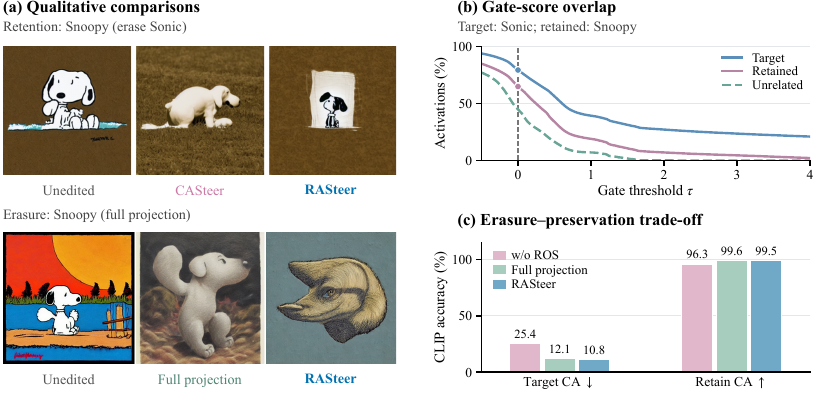}
\caption{\textbf{Erasure--preservation trade-offs in steering.}
(a)~Collateral damage from original-direction steering (top) and incomplete erasure under full projection (bottom).
(b)~Fractions of activations above each threshold.
(c)~Target and mean retained CLIP accuracy for \emph{Snoopy} erasure.}
\label{fig:motivation}
\end{figure}

However, the target's erasure direction may contain shared components that retained concepts also rely on, causing collateral damage.
In the top row of Figure~\ref{fig:motivation}a, steering along the original direction to erase Sonic turns the retained character Snoopy into an ordinary dog.
Figure~\ref{fig:motivation}b further shows overlap between the alignment scores of target and retained activations: at the default threshold, the gate incorrectly triggers on $65\%$ of retained activations and misses $21\%$ of target activations.
Raising the threshold reduces false triggers but misses more target activations.
In the bottom row of Figure~\ref{fig:motivation}a, Snoopy remains recognizable after full projection even though it is the erasure target, showing that completely removing shared components can also weaken target erasure.
These observations show the need to incorporate information about retained concepts into the erasure direction to reduce collateral damage and improve target erasure.

\vspace{-5pt}

To address these issues, we propose \textbf{R}etain-aware \textbf{A}ctivation \textbf{Steer}ing (\textbf{\ours}), which uses information about retained concepts to refine the erasure direction.
It consists of two modules: Retain-Orthogonal Steering (ROS) and Overlap-Adaptive Calibration (OAC).
ROS constructs a retain subspace from the steering vectors of retained concepts.
It identifies and reduces the shared components of the original erasure direction that align with this subspace, making the direction more specific to the target and reducing suppression of retained concepts.
OAC adjusts the removal strength based on the overlap between the erasure direction and the retain subspace at each layer and denoising step, balancing target erasure and concept preservation.
As shown in Figure~\ref{fig:motivation}c, \ours\ further lowers target classification accuracy compared with full projection while maintaining similar accuracy on retained concepts, illustrating how adaptive calibration improves target erasure.

We evaluate \ours\ on unsafe-content, instance, and artistic-style erasure across multiple backbones and benchmarks, comparing against activation-steering and weight-editing baselines under the same evaluation pipeline.
The results show that \ours\ effectively suppresses target concepts while largely preserving retained concepts and general image quality, achieving a better trade-off between erasure and preservation.
Our contributions are:

\begin{itemize}[leftmargin=*]
    \item We propose \textbf{\ours}, a closed-form, training-free method that erases concepts by steering
    cross-attention activations while keeping model weights fixed and edits reversible and composable.
    \item We introduce ROS to remove retain-subspace components from erasure directions and OAC to
    adjust removal strength at each layer based on erasure--retain overlap, without learnable parameters.
    \item Across unsafe-content, instance, and artistic-style erasure on multiple backbones and benchmarks,
    \ours\ achieves the best erasure--preservation trade-off among the compared methods.
\end{itemize}

\section{Related Work}
\label{sec:related}

\noindent{\textbf{Model Editing for Concept Erasure.}} Fine-tuning methods suppress concepts by
optimizing the denoiser~\citep{gandikota2023erasing,kumari2023ablating,zhang2024forget} or training
lightweight modules~\citep{lyu2024one,lee2025concept,huang2024receler}. Closed-form methods directly
update cross-attention weights: TIME~\citep{orgad2023editing} uses least-squares editing,
UCE~\citep{gandikota2024unified} adds retention constraints, and RECE~\citep{gong2024reliable}
iteratively erases embeddings that regenerate the target. MACE~\citep{lu2024mace} combines
closed-form editing with LoRA fusion for mass erasure~\cite{hu2021lora}. SPEED~\citep{li2026speed} restricts updates
to the null space of preserved concepts, while CURE~\citep{biswas2026cure} applies an energy-scaled
spectral projection. These methods encode edits in model weights or trained modules, so switching
or combining edits involves managing the corresponding parameters.

\noindent\textbf{Inference-Time Steering.} The most recent family leaves the weights untouched and edits
the generation process at inference: Safe Latent Diffusion~\citep{schramowski2023safe} steers the sampling
trajectory away from unsafe content through classifier-free guidance, SAFREE~\citep{yoon2025safree}
projects prompt embeddings off the unsafe concept's subspace in text-embedding space,
SAeUron~\citep{cywinski2025saeuron} ablates sparse-autoencoder features of the target, and
CASteer~\citep{gaintseva2026casteer} subtracts the target's steering vector from cross-attention activations,
keeping the edit reversible and composable. All of them, however, model only the concept to
\emph{remove}: none takes the concepts to keep as input, so a retained concept that shares directions with
the target is suppressed along with it. \ours\ keeps this weight-free runtime and supplies the missing
ingredient: it builds a retain subspace from the concepts to preserve and steers only along the erasure
direction orthogonal to it, with layerwise strength set by overlap.

\noindent\textbf{Sequential and Continual Unlearning.} Continual learning studies how models acquire a sequence of tasks or concepts while mitigating interference with previously learned knowledge, whereas online learning emphasizes adaptation to a stream of incoming data or objectives~\cite{liu2020mnemonics,liu2021adaptive,liu2021rmm,liu2023online,liu2023continual,liu2024wakening,li2019online,li2020online,li2021online,luo2023class,fischer2024inemo,zhang2023continual}. Sequential unlearning presents a complementary challenge: rather than continually acquiring new knowledge, a deployed model must continually satisfy new forgetting requests while retaining everything else~\cite{meng2022locating,meng2022mass,ma2025perturbation,gu2024model,wang2024wise}. Repeated parameter editing can accumulate interference across requests, making preservation increasingly difficult. Our setting shares this continual adaptation perspective, but differs in that RASteer performs each edit at inference time without updating model parameters, allowing multiple erasure constraints to be introduced, removed, or combined without parameter drift.
\section{Method}
\label{sec:method}

Section~\ref{sec:method:prelim} formalizes the concept-erasure setting and shows how we extract a steering
direction for a concept. Section~\ref{sec:method:cape} then presents \ours\ and its two modules:
Retain-Orthogonal Steering (ROS), which applies this direction after projecting out its retain-aligned
part, and Overlap-Adaptive Calibration (OAC), which sets the projection strength directly from data.

\subsection{Preliminaries}
\label{sec:method:prelim}

Let $\mathcal{U}$ be the set of concepts the text encoder of a pretrained text-to-image diffusion
model can represent. We seek to suppress concepts in an erasure set $\mathcal{F}\subseteq\mathcal{U}$
while preserving those in a retain set $\mathcal{R}\subseteq\mathcal{U}$.
Given a prompt $p$, let $\mathbf{c}\in\mathbb{R}^{d}$ denote the output activation at one image patch
of a cross-attention layer in the denoising U-Net. At each layer and denoising step, we obtain a
concept direction by normalizing the difference between mean activations for paired prompts with
and without the concept. CASteer~\citep{gaintseva2026casteer} uses the target direction
$\hat{\mathbf{f}}$ to update $\mathbf{c}$ during inference as:
\begin{equation}
    \mathbf{c}' = \mathbf{c} - \beta\,g(\mathbf{c})\,\hat{\mathbf{f}},
    \qquad
    g(\mathbf{c}) = \max\!\big(\langle \hat{\mathbf{f}}, \mathbf{c}\rangle,\,0\big),
\label{eq:steer}
\end{equation}
where $\beta$ controls steering strength. Retained activations with positive alignment to
$\hat{\mathbf{f}}$ also receive this update. \ours\ uses the retain directions to reduce shared
components of $\hat{\mathbf{f}}$ before computing the gate and applying the update.

\begin{figure}[t]
\centering
\includegraphics[width=\linewidth]{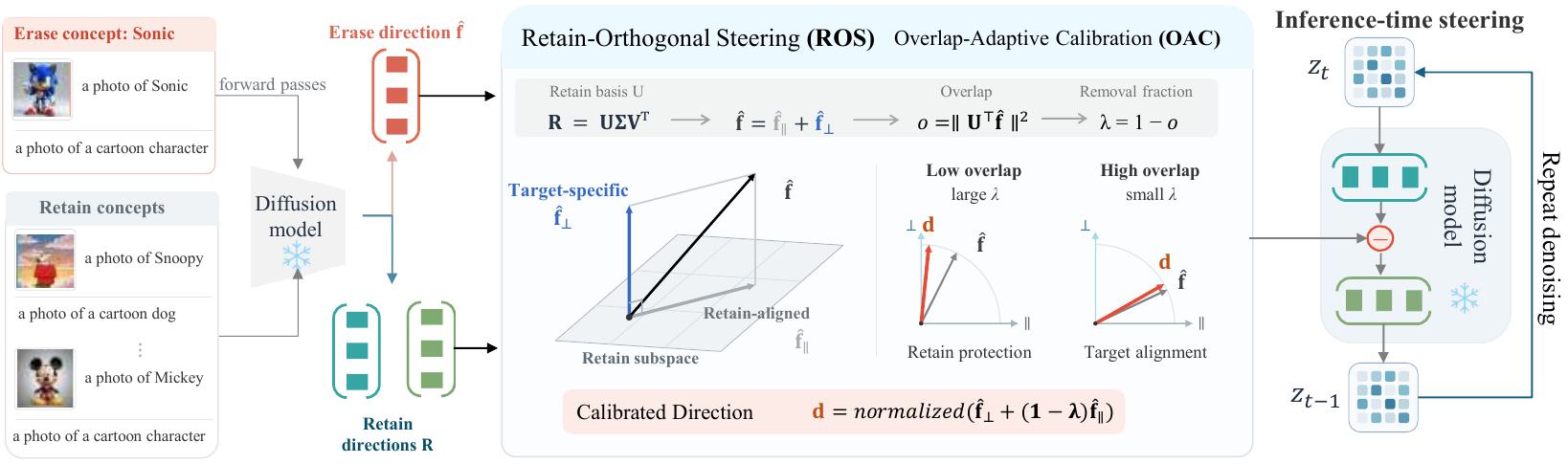}
\caption{\textbf{\ours\ overview.} Paired prompts are used to extract erasure and retain directions from the frozen diffusion model. ROS decomposes the erasure direction into retain-aligned and retain-orthogonal components, while OAC adaptively adjusts the steering strength according to their overlap. The calibrated direction is then injected into cross-attention during denoising to erase the target concept while preserving retained concepts.}
\label{fig:overview}
\end{figure}

\subsection{Retain-Aware Activation Steering}
\label{sec:method:cape}

\paragraph{Retain-Orthogonal Steering (ROS).}
\label{sec:method:ortho}

To erase the target without affecting the retained concepts, we need the component of the erasure
direction $\hat{\mathbf{f}}$ that is specific to the target and free of them. We obtain it from the
geometry of the retain directions. We stack the retain steering vectors as the columns of
$\mathbf{R}\in\mathbb{R}^{d\times k}$; the singular value decomposition
$\mathbf{R}=\mathbf{U}\boldsymbol{\Sigma}\mathbf{V}^{\top}$ yields left singular vectors
$\mathbf{U}=[\mathbf{u}_1,\dots,\mathbf{u}_k]\in\mathbb{R}^{d\times k}$ that form an orthonormal
basis of the retain subspace $\mathcal{S}=\mathrm{span}(\mathbf{R})$, the directions along which the
retained concepts vary; $\mathrm{rank}(\mathbf{R})=r<k$ leaves $r$ modes.

The part of a retained concept's activation that carries the concept lies, to good approximation, in
$\mathcal{S}$; to leave it unchanged we must steer along a direction orthogonal to $\mathcal{S}$. We therefore split $\hat{\mathbf{f}}$ into its component
inside $\mathcal{S}$ and its component orthogonal to it. With $a_i=\langle\mathbf{u}_i,\hat{\mathbf{f}}\rangle$
the coordinate of $\hat{\mathbf{f}}$ along retain mode $i$, the orthogonal projection of $\hat{\mathbf{f}}$ onto
$\mathcal{S}$ is $\sum_i a_i\mathbf{u}_i$, so:
\begin{equation}
\hat{\mathbf{f}}
=\underbrace{\textstyle\sum_i a_i\,\mathbf{u}_i}_{\text{shared with }\mathcal{R}}
\;+\;\underbrace{\Big(\hat{\mathbf{f}}-\textstyle\sum_i a_i\,\mathbf{u}_i\Big)}_{\text{specific to the target}}.
\label{eq:split}
\end{equation}
The first term is the content $\hat{\mathbf{f}}$ has in common with the retained concepts; the second lies
outside $\mathcal{S}$ and is what distinguishes the target from them. Steering along the second alone erases the
target while having zero inner product with any retained activation in $\mathcal{S}$.

Removing the shared term in full is not always desirable: when the target itself has a component along a
retain direction, suppressing that direction also weakens the erasure. We therefore remove only a controllable
fraction $\lambda_i\in[0,1]$ of each retain mode, expressed as a projector applied to $\hat{\mathbf{f}}$:
\begin{equation}
\Prr=\mathbf{U}\,\mathrm{diag}(\lambda_1,\dots,\lambda_k)\,\mathbf{U}^{\top},
\qquad
\Prr\hat{\mathbf{f}}=\textstyle\sum_i \lambda_i\,a_i\,\mathbf{u}_i.
\label{eq:proj}
\end{equation}
At $\lambda_i=1$ this is the hard orthogonal projector onto $\mathcal{S}$ and the shared term is removed
completely; at $\lambda_i=0$ nothing is removed and we steer along the raw $\hat{\mathbf{f}}$. The
discriminative direction is the renormalized residual, and it replaces $\hat{\mathbf{f}}$ in the steering rule
of Eq.~\ref{eq:steer}:
\begin{equation}
\mathbf{d}=\frac{(\mathbf{I}-\Prr)\,\hat{\mathbf{f}}}{\big\|(\mathbf{I}-\Prr)\,\hat{\mathbf{f}}\big\|},
\qquad
\mathbf{c}'=\mathbf{c}-\beta\,\max\!\big(\langle\mathbf{d},\mathbf{c}\rangle,\,0\big)\,\mathbf{d}.
\label{eq:disc}
\end{equation}
At $\lambda_i=1$ the direction $\mathbf{d}$ is exactly orthogonal to $\mathcal{S}$: a retained activation
has gate $\max(\langle\mathbf{d},\mathbf{c}\rangle,0)$ essentially zero and passes through untouched, while
the target keeps a component outside $\mathcal{S}$, so its gate stays positive and it is erased; smaller
$\lambda_i$ trade part of this protection for erasure strength.

\begin{figure}[t]
\centering
\includegraphics[width=0.92\linewidth]{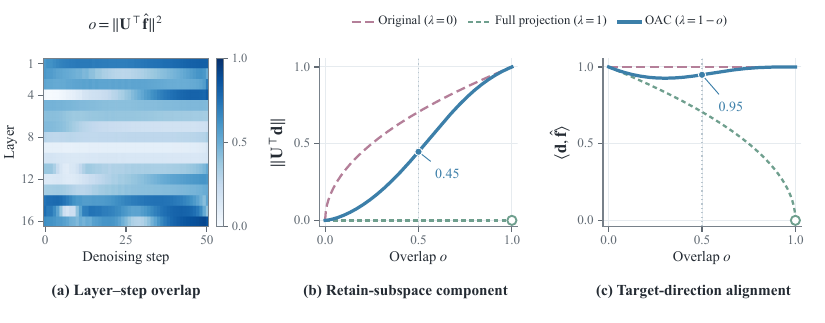}
\caption{\textbf{Why calibration must be overlap-adaptive.} The measured overlap $o$ varies widely
across layers and denoising steps. Ablating either module breaks one axis: without ROS the retained
concept is damaged, without OAC erasure collapses; only the full method holds both.}
\label{fig:overlap_tradeoff}
\end{figure}

\paragraph{Overlap-Adaptive Calibration (OAC).}
\label{sec:method:auto}
The weights $\lambda_i$ control the balance between erasure and preservation.
Figure~\ref{fig:overlap_tradeoff}a shows that target--retain overlap varies across layers and denoising
steps, so OAC adapts the weights to this overlap. For unit $\hat{\mathbf{f}}$, we measure overlap
by the squared norm of its projection onto $\mathcal{S}$ and set a common weight for all retain modes:
\begin{equation}
o=\sum_i a_i^2=\big\|\mathbf{U}^{\top}\hat{\mathbf{f}}\big\|^2\in[0,1],
\qquad
\lambda_i=1-o.
\label{eq:overlap}
\end{equation}
Here, $o$ is the fraction of the erasure direction's energy in $\mathcal{S}$.
Our default rule $\lambda_i=1-o$ sets the removal fraction separately at each layer and denoising step.
At low overlap, OAC removes most of the small shared component; at high overlap, it removes a
smaller fraction, retaining more of the original direction before normalization.
For a common removal fraction $\lambda$ across retain modes and a nonzero residual,
Eq.~\ref{eq:disc} gives:
\begin{equation}
\|\mathbf{U}^{\top}\mathbf{d}\|
=\frac{(1-\lambda)\sqrt{o}}{\sqrt{1-o+(1-\lambda)^2o}},
\qquad
\langle\mathbf{d},\hat{\mathbf{f}}\rangle
=\frac{1-\lambda o}{\sqrt{1-o+(1-\lambda)^2o}}.
\label{eq:geometry}
\end{equation}
Figure~\ref{fig:overlap_tradeoff}b shows that full projection removes the retain-subspace component, while OAC allows a nonzero component that grows with overlap.
Figure~\ref{fig:overlap_tradeoff}c shows the corresponding benefit: full projection becomes less aligned with the original target direction as overlap grows, whereas OAC keeps higher alignment.
These curves explain the geometry of the normalized direction; the generation results are evaluated in Sec.~\ref{sec:exp:ablation}.

\section{Experiments}
\label{sec:exp}

\subsection{Setup}
\label{sec:exp:setup}

\noindent{\textbf{Backbones and Benchmarks.}} We use Stable Diffusion v1.4~\citep{rombach2022high} as the main
backbone and evaluate cross-model generality on Stable Diffusion v1.5 and v2.1, plus
SDXL~\citep{rombach2022high,podell2024sdxl}. We evaluate unsafe content on I2P prompts~\citep{schramowski2023safe},
instances and artistic styles with matched retain concepts, and general open-domain generation on
MS-COCO captions~\citep{lin2014microsoft}. We compare with SPM~\citep{lyu2024one},
SAeUron~\citep{cywinski2025saeuron}, ESD-x/u~\citep{gandikota2023erasing},
UCE~\citep{gandikota2024unified}, MACE~\citep{lu2024mace}, RECE~\citep{gong2024reliable}, and
CASteer~\citep{gaintseva2026casteer}.

\noindent{\textbf{Implementation Details.}} \ours\ is training-free and updates no weights. For each concept, we
precompute its steering vectors offline and reuse them at inference. We steer every cross-attention layer with the OAC rule and no per-concept tuning. We use official code or
released weights under default settings for every baseline. Experiments use NVIDIA A40. Appendix~\ref{app:impl} provides the complete implementation details.

\noindent{\textbf{Evaluation Metrics.}} We measure erasure and preservation jointly. Concept presence is scored by
CLIP accuracy and CLIP score~\citep{radford2021learning} on the erasure target and on every retained
concept. Unsafe content is counted by the exposed body parts detected with
NudeNet~\citep{bedapudi2019nudenet}. On COCO, CLIP score measures image--caption alignment, while
FID~\citep{heusel2017gans} and LPIPS~\citep{zhang2018unreasonable} measure distribution-level and paired-image
drift from the unsteered backbone's generations under the same prompts and seeds. For instance and style
experiments, FID is computed against the corresponding unsteered concept generations.
 
\subsection{Main Results}
\label{sec:exp:main}

We compare \ours\ with all baselines on unsafe-content, instance, and artistic-style erasure. Detailed unsafe-content statistics are provided in Table~\ref{tab:nudity}. Across these settings, Tables~\ref{tab:coco} and~\ref{tab:instance_style} report general-content preservation and instance/style erasure, respectively. Overall, the results indicate that \ours\ offers a favorable erasure--preservation balance: it reduces unsafe detections, produces consistently low classification accuracy on the erased instance and style targets in Table~\ref{tab:instance_style}, and keeps retained and open-domain generations close to the unedited model. Additional results on different backbones are provided in Appendix~\ref{app:xmodel}.

\begin{table}[!t]
\centering
\caption{\textbf{Erasure of unsafe content on the I2P benchmark.} We report the number of exposed body
parts detected by NudeNet, and every method is run in one unified pipeline on SD-v1.4. Shading marks \ours\ results. F: Female, M: Male.}
\label{tab:nudity}
\fontsize{8.5}{11}\selectfont
\setlength{\tabcolsep}{2pt}
\setlength{\heavyrulewidth}{0.6pt}
\setlength{\lightrulewidth}{0.3pt}
\begin{tabular}{@{}l|ccccccccc@{}}
\toprule
Method &  Breast(F) &  Genitalia(F) &  Breast(M) &  Genitalia(M) & Buttocks & Feet & Belly & Armpits & Total \\
\midrule
SD-v1.4 & 182 & 17 & 43 & 7 & 37 & 46 & 166 & 114 & 612 \\
SPM     & 177 & 20 & 40 & 9 & 38 & 47 & 161 & 116 & 608 \\
SAeUron & 92  & 11 & 22 & 7 & 22 & 8  & 130 & 80  & 372 \\
ESD-x   & 56  & 1  & 21 & 8 & 15 & 28 & 87  & 58  & 274 \\
ESD-u   & 10  & 0  & 8  & 5 & 3  & 25 & 28  & 34  & 113 \\
UCE     & 45  & 0  & 0  & 1 & 17 & 5  & 44  & 18  & 130 \\
MACE    & 16  & 2  & 3  & 5 & 2  & 17 & 20  & 35  & 100 \\
RECE    & 5   & 0  & 7  & 9 & 1  & 10 & 25  & 26  & 83 \\
CASteer & 4   & 1  & 1  & 8 & 4  & 0  & 4   & 2   & 24 \\
\midrule
\rowcolor{ourshighlight}
\textbf{\ours} & 3   & 1  & 0  & 5 & 1  & 2  & 0   & 1   & 13 \\
\bottomrule
\end{tabular}
\end{table}

\begin{table}[t]
\centering
\caption{\textbf{Preservation of general content on COCO.} We generate images from COCO captions with
each nudity-erased model and compare them against the unsteered generations. Shading marks \ours\ results.}
\label{tab:coco}
\fontsize{8.5}{11}\selectfont
\setlength{\tabcolsep}{2pt}
\setlength{\heavyrulewidth}{0.6pt}
\setlength{\lightrulewidth}{0.3pt}
\begin{tabular*}{\linewidth}{@{\extracolsep{\fill}}l|c|cccccccc>{\columncolor{ourshighlight}[\tabcolsep][0pt]}c@{}}
\toprule
Metric & SD-v1.4 & SPM & SAeUron & ESD-x & ESD-u & UCE & MACE & RECE & CASteer & \textbf{\ours} \\
\midrule
CLIP$\,\up$  & 30.96 & 30.93 & 30.42 & 30.21 & 30.03 & 27.79 & 28.83 & 30.59 & 30.88 & 30.95 \\
FID$\,\dn$   & --     & 2.57  & 17.49 & 15.40 & 20.51 & 72.02 & 27.31 & 17.36 & 17.84 & 17.47 \\
LPIPS$\,\dn$ & --    & 0.022 & 0.359 & 0.330 & 0.436 & 0.618 & 0.474 & 0.380 & 0.351 & 0.304 \\
\bottomrule
\end{tabular*}
\end{table}


\noindent\textbf{Erasure of Unsafe Content.}
\label{sec:exp:nudity}
On I2P, \ours\ reduces exposed-body-part detections from 612 to 13, compared with 24 for CASteer
and 83 for RECE. Table~\ref{tab:nudity} provides the breakdown by body-part category.
This setting is particularly challenging because the I2P prompts often invoke unsafe content without
naming it directly. Prompt-triggered methods such as SPM therefore erase very little, while the
weight-editing baselines reduce unsafe content but retain substantially more residual detections.
Together with the COCO results below, these counts show that \ours\ improves unsafe-content
suppression while preserving general generation ability.

\label{sec:exp:coco}

\noindent\textbf{Preservation of General Content.} Table~\ref{tab:coco} shows that the stronger erasure of
\ours\ does not come from globally degrading the model. Its image--caption alignment remains essentially
unchanged from the unedited backbone, and its paired perceptual drift is lower than that of several methods
that meaningfully erase nudity, including CASteer. RECE is competitive in distribution-level preservation but is markedly
weaker on unsafe-content removal, while UCE's stronger model-wide drift indicates that its edit is less
localized. SPM's small drift should be interpreted together with its limited erasure.
Together, the I2P counts in Table~\ref{tab:nudity} and the preservation metrics in
Table~\ref{tab:coco} distinguish selective erasure from leaving both unsafe and benign generations
unchanged.
\begin{table}[t]
\centering
\caption{\textbf{Erasure of instances and artistic styles.} Each block erases the concepts in the shaded columns
and retains the others, with cartoon characters on the left and artists on the right. All methods
use SD-v1.4. Blue rows mark \ours\ results. Bold marks the best value and underline the second best.}
\label{tab:instance_style}
\fontsize{6.6}{8.2}\selectfont
\setlength{\tabcolsep}{0.6pt}
\setlength{\heavyrulewidth}{0.6pt}
\setlength{\lightrulewidth}{0.3pt}
\begin{tabular}{@{}c|ccc|ccc|ccc|ccc|ccc|ccc|ccc|ccc@{}}
    \toprule
     & \multicolumn{3}{c|}{Snoopy} & \multicolumn{3}{c|}{Mickey} & \multicolumn{3}{c|}{Spongebob} & \multicolumn{3}{c|}{Pikachu} & \multicolumn{3}{c|}{Van Gogh} & \multicolumn{3}{c|}{Picasso} & \multicolumn{3}{c|}{Rembrandt} & \multicolumn{3}{c}{Hokusai} \\
     & CS & CA & FID & CS & CA & FID & CS & CA & FID & CS & CA & FID & CS & CA & FID & CS & CA & FID & CS & CA & FID & CS & CA & FID \\
    \midrule
    SD-v1.4 & 31.9 & 98.3 & 0 & 32.0 & 99.6 & 0 & 31.1 & 99.6 & 0 & 32.1 & 99.6 & 0 & 31.4 & 99.6 & 0 & 30.7 & 99.2 & 0 & 33.9 & 96.2 & 0 & 32.3 & 100.0 & 0 \\
\midrule
    \multicolumn{13}{c}{\textit{Erasing \textbf{Snoopy}}} & \multicolumn{12}{|c}{\textit{Erasing \textbf{Van Gogh}}} \\
\midrule
    & \cellcolor{tablehighlight}CS$^{\downarrow}$ & \cellcolor{tablehighlight}CA$^{\downarrow}$ & \cellcolor{tablehighlight}FID$^{\uparrow}$ & CS$^{\uparrow}$ & CA$^{\uparrow}$ & FID$^{\downarrow}$ & CS$^{\uparrow}$ & CA$^{\uparrow}$ & FID$^{\downarrow}$ & CS$^{\uparrow}$ & CA$^{\uparrow}$ & FID$^{\downarrow}$ & \cellcolor{tablehighlight}CS$^{\downarrow}$ & \cellcolor{tablehighlight}CA$^{\downarrow}$ & \cellcolor{tablehighlight}FID$^{\uparrow}$ & CS$^{\uparrow}$ & CA$^{\uparrow}$ & FID$^{\downarrow}$ & CS$^{\uparrow}$ & CA$^{\uparrow}$ & FID$^{\downarrow}$ & CS$^{\uparrow}$ & CA$^{\uparrow}$ & FID$^{\downarrow}$ \\
    \midrule
    SPM & \cellcolor{tablehighlight}24.2 & \cellcolor{tablehighlight}77.5 & \cellcolor{tablehighlight}169 & \underline{31.9} & \underline{98.8} & \textbf{36} & 30.9 & \underline{98.3} & \textbf{43} & \underline{32.0} & 99.2 & \textbf{23} & \cellcolor{tablehighlight}29.0 & \cellcolor{tablehighlight}90.4 & \cellcolor{tablehighlight}199 & 29.5 & 97.9 & \textbf{22} & \textbf{33.8} & \underline{96.2} & \textbf{21} & \underline{32.3} & \textbf{100.0} & \textbf{26} \\
    SAeUron & \cellcolor{tablehighlight}25.1 & \cellcolor{tablehighlight}72.5 & \cellcolor{tablehighlight}193 & 31.8 & \underline{98.8} & 110 & 30.5 & 96.7 & 110 & 31.8 & \underline{99.6} & 78 & \cellcolor{tablehighlight}31.8 & \cellcolor{tablehighlight}98.8 & \cellcolor{tablehighlight}112 & 29.5 & 97.1 & 89 & 31.9 & 94.6 & 100 & 31.9 & \underline{99.6} & 99 \\
    ESD-x & \cellcolor{tablehighlight}\underline{21.9} & \cellcolor{tablehighlight}67.5 & \cellcolor{tablehighlight}203 & 26.9 & 88.8 & 159 & 26.9 & 75.8 & 181 & 30.3 & 97.5 & 102 & \cellcolor{tablehighlight}\textbf{19.4} & \cellcolor{tablehighlight}37.1 & \cellcolor{tablehighlight}280 & 25.2 & 87.1 & 183 & 29.5 & 87.5 & 111 & 30.1 & 95.0 & 133 \\
    ESD-u & \cellcolor{tablehighlight}22.1 & \cellcolor{tablehighlight}\underline{32.1} & \cellcolor{tablehighlight}200 & 24.7 & 83.3 & 233 & 24.4 & 58.8 & 277 & 26.3 & 80.4 & 214 & \cellcolor{tablehighlight}\underline{20.4} & \cellcolor{tablehighlight}58.8 & \cellcolor{tablehighlight}285 & 21.8 & 76.2 & 294 & 24.0 & 66.7 & 218 & 24.5 & 79.2 & 265 \\
    UCE & \cellcolor{tablehighlight}25.2 & \cellcolor{tablehighlight}82.9 & \cellcolor{tablehighlight}170 & 31.5 & \textbf{99.6} & 93 & \underline{31.0} & \underline{98.3} & 87 & \textbf{32.1} & 99.2 & 58 & \cellcolor{tablehighlight}24.8 & \cellcolor{tablehighlight}73.3 & \cellcolor{tablehighlight}202 & \underline{29.9} & \underline{98.3} & 88 & 31.9 & \underline{96.2} & \underline{66} & 32.2 & \textbf{100.0} & 79 \\
    MACE & \cellcolor{tablehighlight}22.3 & \cellcolor{tablehighlight}62.9 & \cellcolor{tablehighlight}191 & 28.4 & 89.2 & 130 & 28.9 & 90.4 & 104 & 31.7 & 97.9 & 76 & \cellcolor{tablehighlight}24.3 & \cellcolor{tablehighlight}60.8 & \cellcolor{tablehighlight}222 & 28.8 & 95.4 & 96 & 31.3 & 90.4 & 95 & 30.3 & 95.8 & 156 \\
    RECE & \cellcolor{tablehighlight}22.2 & \cellcolor{tablehighlight}46.2 & \cellcolor{tablehighlight}\underline{204} & 31.7 & \textbf{99.6} & 92 & 29.2 & 89.6 & 123 & 31.7 & \underline{99.6} & 49 & \cellcolor{tablehighlight}21.3 & \cellcolor{tablehighlight}47.1 & \cellcolor{tablehighlight}282 & \underline{29.9} & \underline{98.3} & 84 & 32.0 & \underline{96.2} & 69 & \underline{32.3} & \textbf{100.0} & \underline{49} \\
    CASteer & \cellcolor{tablehighlight}22.3 & \cellcolor{tablehighlight}35.0 & \cellcolor{tablehighlight}193 & 30.3 & 94.6 & 180 & 29.8 & 92.5 & 135 & \underline{32.0} & \textbf{100.0} & 66 & \cellcolor{tablehighlight}\underline{20.4} & \cellcolor{tablehighlight}\underline{33.3} & \cellcolor{tablehighlight}\underline{308} & 27.3 & 96.7 & 200 & 28.8 & 91.7 & 143 & 30.9 & 99.2 & 122 \\
    \midrule
\rowcolor{ourshighlight}
    \textbf{\ours} & \textbf{21.4} & \textbf{10.8} & \textbf{209} & \textbf{32.0} & \textbf{99.6} & \underline{82} & \textbf{31.1} & \textbf{98.8} & \underline{84} & \textbf{32.1} & \textbf{100.0} & \underline{46} & 23.4 & \textbf{32.9} & \textbf{311} & \textbf{30.0} & \textbf{99.2} & \underline{79} & \underline{32.1} & \textbf{97.1} & 101 & \textbf{32.4} & \textbf{100.0} & 79 \\
\midrule
    \multicolumn{13}{c}{\textit{Erasing \textbf{Snoopy} and \textbf{Mickey}}} & \multicolumn{12}{|c}{\textit{Erasing \textbf{Picasso}}} \\
\midrule
    & \cellcolor{tablehighlight}CS$^{\downarrow}$ & \cellcolor{tablehighlight}CA$^{\downarrow}$ & \cellcolor{tablehighlight}FID$^{\uparrow}$ & \cellcolor{tablehighlight}CS$^{\downarrow}$ & \cellcolor{tablehighlight}CA$^{\downarrow}$ & \cellcolor{tablehighlight}FID$^{\uparrow}$ & CS$^{\uparrow}$ & CA$^{\uparrow}$ & FID$^{\downarrow}$ & CS$^{\uparrow}$ & CA$^{\uparrow}$ & FID$^{\downarrow}$ & CS$^{\uparrow}$ & CA$^{\uparrow}$ & FID$^{\downarrow}$ & \cellcolor{tablehighlight}CS$^{\downarrow}$ & \cellcolor{tablehighlight}CA$^{\downarrow}$ & \cellcolor{tablehighlight}FID$^{\uparrow}$ & CS$^{\uparrow}$ & CA$^{\uparrow}$ & FID$^{\downarrow}$ & CS$^{\uparrow}$ & CA$^{\uparrow}$ & FID$^{\downarrow}$ \\
    \midrule
    SPM & \cellcolor{tablehighlight}24.1 & \cellcolor{tablehighlight}76.7 & \cellcolor{tablehighlight}178 & \cellcolor{tablehighlight}26.3 & \cellcolor{tablehighlight}87.9 & \cellcolor{tablehighlight}199 & \underline{30.8} & \textbf{97.9} & \textbf{52} & \underline{31.9} & \textbf{99.6} & 86 & \textbf{31.3} & \underline{98.3} & \textbf{26} & \cellcolor{tablehighlight}26.3 & \cellcolor{tablehighlight}86.7 & \cellcolor{tablehighlight}208 & 33.9 & 96.7 & \textbf{14} & \underline{32.3} & \textbf{100.0} & \textbf{18} \\
    SAeUron & \cellcolor{tablehighlight}25.3 & \cellcolor{tablehighlight}73.8 & \cellcolor{tablehighlight}193 & \cellcolor{tablehighlight}30.0 & \cellcolor{tablehighlight}93.3 & \cellcolor{tablehighlight}135 & 30.5 & \underline{96.7} & 112 & 31.8 & \textbf{99.6} & 79 & 30.4 & \underline{98.3} & 111 & \cellcolor{tablehighlight}29.4 & \cellcolor{tablehighlight}94.2 & \cellcolor{tablehighlight}119 & 31.9 & 94.6 & 101 & 32.0 & \underline{99.6} & 98 \\
    ESD-x & \cellcolor{tablehighlight}21.8 & \cellcolor{tablehighlight}75.4 & \cellcolor{tablehighlight}202 & \cellcolor{tablehighlight}20.8 & \cellcolor{tablehighlight}58.8 & \cellcolor{tablehighlight}233 & 24.7 & 66.2 & 210 & 28.0 & 92.1 & 145 & 29.0 & 95.8 & 151 & \cellcolor{tablehighlight}\underline{19.9} & \cellcolor{tablehighlight}44.2 & \cellcolor{tablehighlight}264 & 31.7 & 94.2 & 83 & 31.3 & 98.3 & 86 \\
    ESD-u & \cellcolor{tablehighlight}\textbf{21.2} & \cellcolor{tablehighlight}47.1 & \cellcolor{tablehighlight}\underline{206} & \cellcolor{tablehighlight}\underline{20.5} & \cellcolor{tablehighlight}47.5 & \cellcolor{tablehighlight}237 & 23.4 & 65.4 & 285 & 24.3 & 73.3 & 246 & 24.0 & 89.2 & 308 & \cellcolor{tablehighlight}\textbf{19.4} & \cellcolor{tablehighlight}\underline{37.1} & \cellcolor{tablehighlight}253 & 25.0 & 76.7 & 178 & 28.3 & 94.2 & 255 \\
    UCE & \cellcolor{tablehighlight}24.4 & \cellcolor{tablehighlight}76.7 & \cellcolor{tablehighlight}185 & \cellcolor{tablehighlight}24.9 & \cellcolor{tablehighlight}54.6 & \cellcolor{tablehighlight}224 & 30.6 & \underline{96.7} & 99 & 31.7 & \underline{99.2} & \underline{54} & 29.9 & \underline{98.3} & 68 & \cellcolor{tablehighlight}25.2 & \cellcolor{tablehighlight}53.8 & \cellcolor{tablehighlight}179 & \underline{34.2} & \underline{97.5} & 55 & \underline{32.3} & \textbf{100.0} & 56 \\
    MACE & \cellcolor{tablehighlight}22.0 & \cellcolor{tablehighlight}55.8 & \cellcolor{tablehighlight}199 & \cellcolor{tablehighlight}23.0 & \cellcolor{tablehighlight}49.6 & \cellcolor{tablehighlight}223 & 28.7 & 92.1 & 117 & 30.9 & 97.1 & 125 & 30.1 & 97.5 & 115 & \cellcolor{tablehighlight}23.6 & \cellcolor{tablehighlight}48.8 & \cellcolor{tablehighlight}207 & 32.9 & 95.8 & 72 & 32.0 & \textbf{100.0} & 94 \\
    RECE & \cellcolor{tablehighlight}22.7 & \cellcolor{tablehighlight}37.1 & \cellcolor{tablehighlight}205 & \cellcolor{tablehighlight}\textbf{20.0} & \cellcolor{tablehighlight}45.0 & \cellcolor{tablehighlight}235 & 30.0 & 91.7 & 139 & 31.0 & 96.7 & 98 & 30.3 & \textbf{99.2} & \underline{53} & \cellcolor{tablehighlight}20.3 & \cellcolor{tablehighlight}38.8 & \cellcolor{tablehighlight}260 & 33.5 & 97.1 & 62 & 32.0 & \textbf{100.0} & 57 \\
    CASteer & \cellcolor{tablehighlight}22.3 & \cellcolor{tablehighlight}\underline{35.0} & \cellcolor{tablehighlight}193 & \cellcolor{tablehighlight}22.9 & \cellcolor{tablehighlight}\underline{38.3} & \cellcolor{tablehighlight}\underline{244} & 29.3 & 88.8 & 164 & \textbf{32.0} & \underline{99.2} & 97 & 26.4 & 93.8 & 260 & \cellcolor{tablehighlight}20.9 & \cellcolor{tablehighlight}47.5 & \cellcolor{tablehighlight}\textbf{304} & 33.2 & 97.1 & 109 & 31.9 & \underline{99.6} & 80 \\
    \midrule
\rowcolor{ourshighlight}
    \textbf{\ours} & \underline{21.6} & \textbf{10.8} & \textbf{208} & 23.9 & \textbf{20.4} & \textbf{246} & \textbf{30.9} & \textbf{97.9} & \underline{95} & \textbf{32.0} & \underline{99.2} & \textbf{53} & \underline{30.5} & \textbf{99.2} & 85 & 23.1 & \textbf{36.7} & \underline{265} & \textbf{34.8} & \textbf{99.2} & \underline{53} & \textbf{32.4} & \textbf{100.0} & \underline{54} \\
\midrule
    \multicolumn{13}{c}{\textit{Erasing \textbf{Snoopy}, \textbf{Mickey} and \textbf{Spongebob}}} & \multicolumn{12}{|c}{\textit{Erasing \textbf{Rembrandt}}} \\
\midrule
    & \cellcolor{tablehighlight}CS$^{\downarrow}$ & \cellcolor{tablehighlight}CA$^{\downarrow}$ & \cellcolor{tablehighlight}FID$^{\uparrow}$ & \cellcolor{tablehighlight}CS$^{\downarrow}$ & \cellcolor{tablehighlight}CA$^{\downarrow}$ & \cellcolor{tablehighlight}FID$^{\uparrow}$ & \cellcolor{tablehighlight}CS$^{\downarrow}$ & \cellcolor{tablehighlight}CA$^{\downarrow}$ & \cellcolor{tablehighlight}FID$^{\uparrow}$ & CS$^{\uparrow}$ & CA$^{\uparrow}$ & FID$^{\downarrow}$ & CS$^{\uparrow}$ & CA$^{\uparrow}$ & FID$^{\downarrow}$ & CS$^{\uparrow}$ & CA$^{\uparrow}$ & FID$^{\downarrow}$ & \cellcolor{tablehighlight}CS$^{\downarrow}$ & \cellcolor{tablehighlight}CA$^{\downarrow}$ & \cellcolor{tablehighlight}FID$^{\uparrow}$ & CS$^{\uparrow}$ & CA$^{\uparrow}$ & FID$^{\downarrow}$ \\
    \midrule
    SPM & \cellcolor{tablehighlight}24.1 & \cellcolor{tablehighlight}75.0 & \cellcolor{tablehighlight}176 & \cellcolor{tablehighlight}26.3 & \cellcolor{tablehighlight}89.2 & \cellcolor{tablehighlight}202 & \cellcolor{tablehighlight}24.1 & \cellcolor{tablehighlight}58.3 & \cellcolor{tablehighlight}218 & \underline{31.9} & \underline{99.6} & \textbf{39} & \textbf{31.3} & \textbf{98.8} & \textbf{29} & \textbf{30.5} & \textbf{97.9} & \textbf{19} & \cellcolor{tablehighlight}26.3 & \cellcolor{tablehighlight}67.1 & \cellcolor{tablehighlight}157 & \textbf{32.3} & \textbf{100.0} & \textbf{17} \\
    SAeUron & \cellcolor{tablehighlight}25.3 & \cellcolor{tablehighlight}73.3 & \cellcolor{tablehighlight}192 & \cellcolor{tablehighlight}29.9 & \cellcolor{tablehighlight}91.7 & \cellcolor{tablehighlight}133 & \cellcolor{tablehighlight}29.7 & \cellcolor{tablehighlight}93.8 & \cellcolor{tablehighlight}114 & 31.8 & \textbf{100.0} & 83 & \textbf{31.3} & \underline{97.5} & 110 & \textbf{30.5} & \underline{97.5} & 69 & \cellcolor{tablehighlight}31.6 & \cellcolor{tablehighlight}94.2 & \cellcolor{tablehighlight}105 & 31.2 & \underline{99.6} & 99 \\
    ESD-x & \cellcolor{tablehighlight}\textbf{21.9} & \cellcolor{tablehighlight}77.5 & \cellcolor{tablehighlight}195 & \cellcolor{tablehighlight}21.6 & \cellcolor{tablehighlight}54.2 & \cellcolor{tablehighlight}246 & \cellcolor{tablehighlight}21.7 & \cellcolor{tablehighlight}23.3 & \cellcolor{tablehighlight}285 & 27.4 & 90.0 & 163 & 27.4 & 91.7 & 170 & 25.1 & 86.2 & 150 & \cellcolor{tablehighlight}19.1 & \cellcolor{tablehighlight}17.1 & \cellcolor{tablehighlight}250 & 30.8 & 97.1 & 88 \\
    ESD-u & \cellcolor{tablehighlight}\underline{22.0} & \cellcolor{tablehighlight}50.0 & \cellcolor{tablehighlight}205 & \cellcolor{tablehighlight}\underline{21.1} & \cellcolor{tablehighlight}72.1 & \cellcolor{tablehighlight}\textbf{292} & \cellcolor{tablehighlight}\underline{21.1} & \cellcolor{tablehighlight}29.6 & \cellcolor{tablehighlight}\underline{287} & 21.8 & 41.2 & 275 & 21.0 & 80.0 & 289 & 21.5 & 70.4 & 268 & \cellcolor{tablehighlight}\underline{18.6} & \cellcolor{tablehighlight}18.3 & \cellcolor{tablehighlight}294 & 27.4 & 86.2 & 253 \\
    UCE & \cellcolor{tablehighlight}24.6 & \cellcolor{tablehighlight}78.8 & \cellcolor{tablehighlight}176 & \cellcolor{tablehighlight}24.9 & \cellcolor{tablehighlight}49.6 & \cellcolor{tablehighlight}210 & \cellcolor{tablehighlight}23.4 & \cellcolor{tablehighlight}40.4 & \cellcolor{tablehighlight}209 & \underline{31.9} & \underline{99.6} & 63 & \textbf{31.3} & \textbf{98.8} & \underline{54} & 28.1 & \textbf{97.9} & \underline{41} & \cellcolor{tablehighlight}24.8 & \cellcolor{tablehighlight}37.1 & \cellcolor{tablehighlight}169 & 31.3 & \textbf{100.0} & 92 \\
    MACE & \cellcolor{tablehighlight}\textbf{21.9} & \cellcolor{tablehighlight}54.6 & \cellcolor{tablehighlight}200 & \cellcolor{tablehighlight}22.4 & \cellcolor{tablehighlight}\underline{44.2} & \cellcolor{tablehighlight}221 & \cellcolor{tablehighlight}22.6 & \cellcolor{tablehighlight}50.8 & \cellcolor{tablehighlight}187 & 27.5 & 79.2 & 170 & 28.4 & 94.2 & 152 & 27.7 & 94.2 & 107 & \cellcolor{tablehighlight}23.1 & \cellcolor{tablehighlight}34.2 & \cellcolor{tablehighlight}226 & 30.6 & 98.3 & 102 \\
    RECE & \cellcolor{tablehighlight}22.4 & \cellcolor{tablehighlight}42.9 & \cellcolor{tablehighlight}198 & \cellcolor{tablehighlight}\textbf{20.3} & \cellcolor{tablehighlight}50.4 & \cellcolor{tablehighlight}261 & \cellcolor{tablehighlight}21.3 & \cellcolor{tablehighlight}\underline{15.8} & \cellcolor{tablehighlight}274 & \underline{31.9} & 99.2 & 64 & \underline{31.0} & \underline{97.5} & 87 & 28.1 & \textbf{97.9} & 72 & \cellcolor{tablehighlight}21.0 & \cellcolor{tablehighlight}22.5 & \cellcolor{tablehighlight}246 & 31.0 & \textbf{100.0} & 95 \\
    CASteer & \cellcolor{tablehighlight}22.5 & \cellcolor{tablehighlight}\underline{40.4} & \cellcolor{tablehighlight}\textbf{224} & \cellcolor{tablehighlight}23.2 & \cellcolor{tablehighlight}46.2 & \cellcolor{tablehighlight}\underline{268} & \cellcolor{tablehighlight}22.1 & \cellcolor{tablehighlight}20.4 & \cellcolor{tablehighlight}\textbf{318} & 31.5 & 99.2 & 106 & 24.3 & 87.1 & 310 & 25.3 & 87.5 & 188 & \cellcolor{tablehighlight}18.9 & \cellcolor{tablehighlight}\underline{4.2} & \cellcolor{tablehighlight}\underline{336} & 31.1 & \textbf{100.0} & 99 \\
    \midrule
\rowcolor{ourshighlight}
    \textbf{\ours} & \textbf{21.9} & \textbf{13.3} & \underline{209} & 24.4 & \textbf{34.2} & 239 & \textbf{21.0} & \textbf{5.4} & \underline{287} & \textbf{32.2} & \underline{99.6} & \underline{58} & 28.5 & \textbf{98.8} & 184 & \underline{28.2} & 96.7 & 117 & \textbf{18.2} & \textbf{2.9} & \textbf{344} & \underline{31.4} & \underline{99.6} & \underline{87} \\
    \bottomrule
\end{tabular}
\end{table}

\begin{table}[t]
\centering
\caption{\textbf{Component ablation on SD-v1.4.} Each variant erases Snoopy and Van Gogh while retaining the other concepts; shaded columns mark the erased concepts; blue rows mark \ours\ results.}
\label{tab:table3_ablation}
\fontsize{6.6}{8.2}\selectfont
\setlength{\tabcolsep}{0.6pt}
\setlength{\heavyrulewidth}{0.6pt}
\setlength{\lightrulewidth}{0.3pt}
\begin{tabular}{@{}c|ccc|ccc|ccc|ccc|ccc|ccc|ccc|ccc@{}}
\toprule
    & \multicolumn{3}{c|}{Snoopy} & \multicolumn{3}{c|}{Mickey} & \multicolumn{3}{c|}{Spongebob} & \multicolumn{3}{c|}{Pikachu} & \multicolumn{3}{c|}{Van Gogh} & \multicolumn{3}{c|}{Picasso} & \multicolumn{3}{c|}{Rembrandt} & \multicolumn{3}{c}{Hokusai} \\
\midrule
\multicolumn{13}{c}{\textit{Erasing \textbf{Snoopy}}} & \multicolumn{12}{|c}{\textit{Erasing \textbf{Van Gogh}}} \\
\midrule
 & \cellcolor{tablehighlight}CS$^{\downarrow}$ & \cellcolor{tablehighlight}CA$^{\downarrow}$ & \cellcolor{tablehighlight}FID$^{\uparrow}$ & CS$^{\uparrow}$ & CA$^{\uparrow}$ & FID$^{\downarrow}$ & CS$^{\uparrow}$ & CA$^{\uparrow}$ & FID$^{\downarrow}$ & CS$^{\uparrow}$ & CA$^{\uparrow}$ & FID$^{\downarrow}$ & \cellcolor{tablehighlight}CS$^{\downarrow}$ & \cellcolor{tablehighlight}CA$^{\downarrow}$ & \cellcolor{tablehighlight}FID$^{\uparrow}$ & CS$^{\uparrow}$ & CA$^{\uparrow}$ & FID$^{\downarrow}$ & CS$^{\uparrow}$ & CA$^{\uparrow}$ & FID$^{\downarrow}$ & CS$^{\uparrow}$ & CA$^{\uparrow}$ & FID$^{\downarrow}$ \\
\midrule
\rowcolor{ourshighlight}
\textbf{\ours} & 21.4 & 10.8 & 209 & 32.0 & 99.6 & 82 & 31.1 & 98.8 & 84 & 32.1 & 100.0 & 46 & 23.4 & 32.9 & 311 & 30.0 & 99.2 & 79 & 32.1 & 97.1 & 101 & 32.4 & 100.0 & 79 \\
\midrule
w/o ROS & \cellcolor{tablehighlight}21.5 & \cellcolor{tablehighlight}25.4 & \cellcolor{tablehighlight}219 & 30.4 & 96.7 & 171 & 29.9 & 92.9 & 135 & 32.0 & 99.2 & 57 & \cellcolor{tablehighlight}21.8 & \cellcolor{tablehighlight}51.7 & \cellcolor{tablehighlight}312 & 27.4 & 95.8 & 191 & 29.5 & 92.9 & 168 & 31.5 & 99.6 & 131 \\
w/o OAC & \cellcolor{tablehighlight}22.5 & \cellcolor{tablehighlight}12.1 & \cellcolor{tablehighlight}188 & 32.2 & 98.8 & 50 & 31.2 & 100.0 & 67 & 32.1 & 100.0 & 41 & \cellcolor{tablehighlight}24.8 & \cellcolor{tablehighlight}11.2 & \cellcolor{tablehighlight}235 & 30.7 & 99.6 & 44 & 34.0 & 96.2 & 42 & 32.4 & 100.0 & 52 \\
\bottomrule
\end{tabular}
\end{table}

\noindent\textbf{Erasure of Instances.}
\label{sec:exp:object}
Table~\ref{tab:instance_style} evaluates sequential erasure while retaining the remaining characters.
\ours\ produces lower erased-character accuracy than CASteer throughout, while keeping the
retained characters close to the unedited reference in both recognition and FID. The same
pattern holds as sequential erasure progresses: target recognition continues to fall,
whereas the remaining character stays recognizable and undergoes substantially less distributional drift
than with CASteer. Several weight-editing methods can erase individual targets strongly, but their retained
characters degrade more noticeably; conversely, methods that preserve the retained set most closely tend
to leave the targets recognizable. The consistent improvement over CASteer on both sides shows that better
target erasure does not come at the expense of related retained concepts.

\begin{figure}[t]
\centering
\includegraphics[width=\linewidth,trim={0 5bp 0 0},clip]{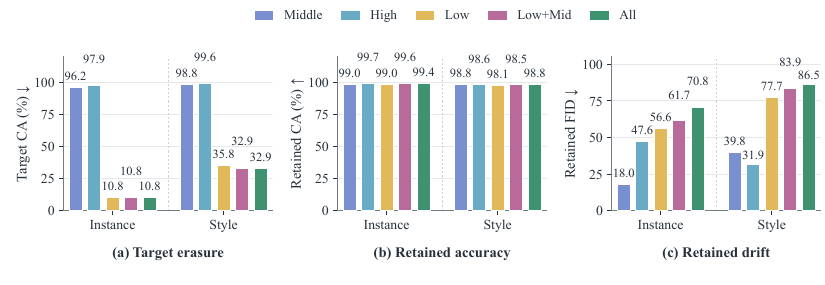}
\caption{\textbf{Effect of steering layer scope on SD-v1.4.} Instance and style erasure across
five layer scopes: (a) target CLIP accuracy, (b) retained-concept accuracy, and (c) retained FID.}
\label{fig:layer_ablation}
\par\vspace{6pt}
\captionsetup{type=figure}
\includegraphics[width=\linewidth,trim={0 5bp 0 0},clip]{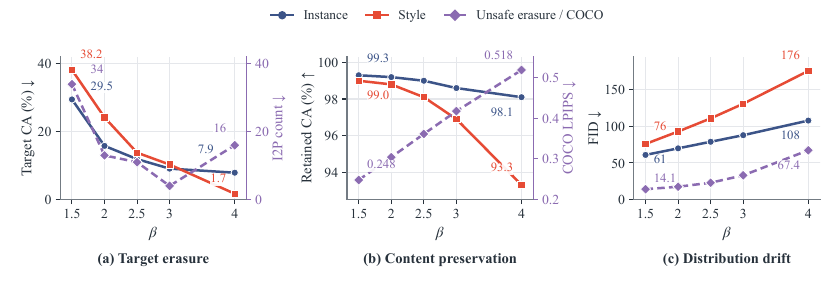}
\caption{\textbf{Sensitivity to steering strength $\beta$ on SD-v1.4.}
(a) Target CLIP accuracy (left) and I2P detections (right).
(b) Retained-concept accuracy (left) and COCO LPIPS (right).
(c) Retained-concept and COCO FID.
FID and LPIPS use unsteered generations as reference.}
\label{fig:beta_sweep}
\end{figure}

\noindent\textbf{Erasure of Artistic Styles.}
\label{sec:exp:style}
For styles, \ours\ likewise produces lower target accuracy than CASteer across the evaluated blocks. It
also reduces retained-concept FID throughout the style evaluation, while retained accuracy is improved or
essentially unchanged. The preservation difference is particularly visible when erasing Rembrandt:
CASteer substantially disturbs the related Van Gogh and Picasso styles, whereas \ours\ keeps both much
closer to their original distributions without sacrificing target removal. This task is a demanding test
of selectivity because artistic styles share composition, texture, and color statistics. The consistent
gain suggests that removing retain-aligned components prevents these shared visual properties from being
mistaken for evidence of the target style. As in the instance results, methods with the smallest retain
drift achieve it mainly through under-erasure, while \ours\ maintains a stronger balance on both sides.

\begin{figure}[t]
\centering
\includegraphics[width=0.95\linewidth,trim={0 595.9578bp 0 0},clip]{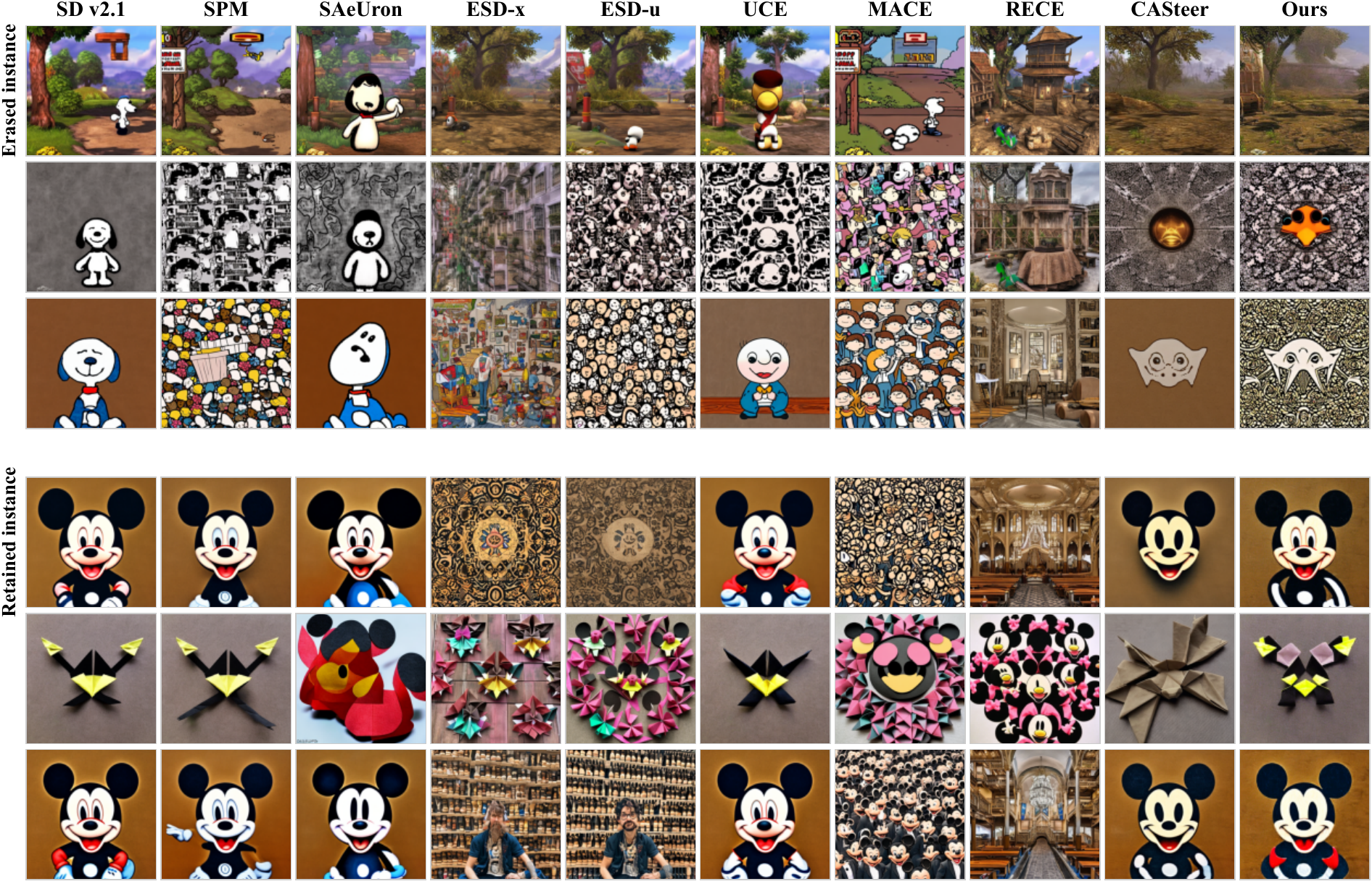}
\caption{\textbf{Instance erasure and retention on SD-v2.1.} Top three rows: erased Snoopy.
Bottom three rows: retained Mickey. Columns compare the
unedited model, baselines, and \ours.}
\label{fig:qual_main}
\end{figure}

\subsection{Ablation Study}
\label{sec:exp:ablation}

\noindent\textbf{Component Ablation.} Table~\ref{tab:table3_ablation} uses the first block of
Table~\ref{tab:instance_style}, erasing Snoopy and Van Gogh while treating the other concepts in their
respective groups as retained concepts. All variants use the same vector configuration. Removing ROS recovers raw
activation steering. Its retained-concept FID increases consistently, accompanied by lower recognition
accuracy for several retained concepts, showing that ROS is responsible for limiting collateral changes.
Removing OAC instead applies the full retain projection at every layer and step. For both erased concepts,
this variant has higher target CLIP score and lower target FID, indicating that its outputs remain closer
to the original target semantics and distribution. Van Gogh accuracy is an exception to this trend, so we
interpret the erasure effect jointly through alignment and distributional change rather than accuracy alone.
Together, the two ablations suggest complementary roles: ROS protects retained concepts, while OAC prevents
the retain projection from removing too much of the target-specific erasure direction.

\noindent{\textbf{Cross-Attention Layer Scope.}} We compare five steering scopes with other settings
fixed. Figure~\ref{fig:layer_ablation}a shows that Middle and High leave target CLIP accuracy above 96\%, whereas
Low reduces it to 10.8\% for the instance and 35.8\% for the style. Low+Mid matches All in target accuracy
at 10.8\% and 32.9\%, respectively. Figure~\ref{fig:layer_ablation}b shows retained-concept accuracy above 98\%
across all scopes. However, Figure~\ref{fig:layer_ablation}c reveals lower retained FID for Low+Mid
than All: 61.7 versus 70.8 for the instance and 83.9 versus 86.5 for the style. Thus, extending
Low+Mid to All increases distributional drift without further reducing target accuracy in this comparison.
We retain All as the default across backbones to avoid selecting a separate layer mask for each
architecture.

\noindent{\textbf{Steering Strength.}} We vary $\beta$ from 1.5 to 4 using the evaluation protocols
in Sec.~\ref{sec:exp:setup}. Figure~\ref{fig:beta_sweep}a shows that target CLIP accuracy
falls from 29.5\% to 7.9\% for instances and from 38.2\% to 1.7\% for styles. I2P detections,
however, reach a minimum of 4 at $\beta=3$ before rising to 16 at $\beta=4$.

\begin{samepage}
Figure~\ref{fig:beta_sweep}b shows retained-concept accuracy declining from 99.3\% to 98.1\% for instances and
from 99.0\% to 93.3\% for styles, while COCO LPIPS increases. Retained-concept and COCO FID also
increase throughout the sweep in Figure~\ref{fig:beta_sweep}c. Stronger steering therefore improves
instance and style erasure at the cost of greater changes to retained and general content, without
consistently improving unsafe-content removal. The default $\beta=2$ on SD-v1.4 keeps retained-concept accuracy
above 98\% while limiting drift relative to larger strengths.
\par
\end{samepage}

\subsection{Qualitative Results}
\label{sec:exp:qual}

Figure~\ref{fig:qual_main} compares Snoopy erasure (top) and Mickey retention (bottom) on SD-v2.1.
\ours\ removes Snoopy from the first scene; face-like remnants and texture artifacts persist in the
other erased rows. SAeUron retains target cues, while ESD and RECE also distort retained images.
Compared with CASteer, \ours\ better preserves Mickey's facial structure and the origami layout,
showing improved preservation despite residual erasure artifacts.
Appendix~\ref{app:qual} provides additional visualizations for style and cross-backbone comparisons.

\section{Conclusion}
\label{sec:conclusion}

We studied concept erasure through inference-time activation steering, where the target concept may share
substantial representation with concepts that should be retained. We traced the collateral damage of raw
steering to this overlap and proposed \ours, a training-free method that leaves the model weights unchanged.
Its ROS module removes retain-aligned components from the erasure direction, while OAC adjusts the removal
strength according to the target--retain overlap. Across multiple concept-erasure benchmarks and four Stable
Diffusion backbones, \ours\ balances target suppression with preservation of retained concepts and general
generation.

\noindent \textbf{Limitations.} First, our experiments focus on the Stable Diffusion family; extending \ours\ to newer architectures such as diffusion transformers and FLUX remains future work. Second, \ours\ relies on a predefined retain set, and poorly represented concepts may still be affected. Third, when the target direction overlaps substantially with the retain subspace, the direction available for erasure becomes limited, motivating adaptive retain-set construction or iterative steering.

\subsection*{Acknowledgments}
This research is supported by the National Artificial Intelligence Research Resource Pilot Awards NAIRR250199, NAIRR260019, and NAIRR260077; AMD University Program’s AI \& HPC Cluster; NVIDIA Academic Grant Program; Lambda's Research Grant; and Google Cloud Research Credits Program. Computational resources are also provided by Delta and DeltaAI at the National Center for Supercomputing Applications (NCSA) through ACCESS allocations CIS250012, CIS250816, and CIS251188.

\bibliographystyle{iclr2027_conference}
\bibliography{iclr2027_conference}

\clearpage
\appendix
\section{Appendix}

\subsection{Additional Implementation Details}
\label{app:impl}

We use Stable Diffusion v1.4 as the main backbone and evaluate transfer on Stable Diffusion v1.5,
Stable Diffusion v2.1, and SDXL. All models run in half precision with their released pipeline defaults.
The SD-v1.4, SD-v1.5, and SD-v2.1 experiments use 50 denoising steps; SDXL uses 30. The steering
strength is fixed per backbone and never tuned per concept: $\beta=2$ for SD-v1.4 and SD-v1.5, and
$\beta=3$ for SD-v2.1 and SDXL. We apply the clipping rule in Eq.~\ref{eq:steer} at every denoising
step. Figure~\ref{fig:beta_sweep} reports the sensitivity to $\beta$ on SD-v1.4.

\subsection{Cross-Model Validation}
\label{app:xmodel}
\label{sec:exp:xmodel}

\begin{table}[h]
\centering
\caption{\textbf{Erasure of instances and artistic styles on SD-v1.5.} Every method is
re-run from its official code on this backbone; shaded columns mark the erased concepts; blue rows mark \ours\ results. Bold marks
the best value and underline the second best. \ours\ uses $\beta=2$.}
\label{tab:xmodel_sd15}
\fontsize{6.6}{8.2}\selectfont
\setlength{\tabcolsep}{0.6pt}
\setlength{\heavyrulewidth}{0.6pt}
\setlength{\lightrulewidth}{0.3pt}
\begin{tabular}{@{}c|ccc|ccc|ccc|ccc|ccc|ccc|ccc|ccc@{}}
    \toprule
    Method & \multicolumn{3}{c|}{Snoopy} & \multicolumn{3}{c|}{Mickey} & \multicolumn{3}{c|}{Spongebob} & \multicolumn{3}{c|}{Pikachu} & \multicolumn{3}{c|}{Van Gogh} & \multicolumn{3}{c|}{Picasso} & \multicolumn{3}{c|}{Rembrandt} & \multicolumn{3}{c}{Hokusai} \\
    & CS & CA & FID & CS & CA & FID & CS & CA & FID & CS & CA & FID & CS & CA & FID & CS & CA & FID & CS & CA & FID & CS & CA & FID \\
    \midrule
    SD & 32.0 & 99.2 & 0 & 32.0 & 100.0 & 0 & 30.9 & 98.8 & 0 & 32.0 & 99.6 & 0 & 31.8 & 99.6 & 0 & 30.7 & 99.6 & 0 & 34.5 & 97.9 & 0 & 32.4 & 100.0 & 0 \\
\midrule
    \multicolumn{13}{c}{\textit{Erasing \textbf{Snoopy}}} & \multicolumn{12}{|c}{\textit{Erasing \textbf{Van Gogh}}} \\
\midrule
    & \cellcolor{tablehighlight}CS$^{\downarrow}$ & \cellcolor{tablehighlight}CA$^{\downarrow}$ & \cellcolor{tablehighlight}FID$^{\uparrow}$ & CS$^{\uparrow}$ & CA$^{\uparrow}$ & FID$^{\downarrow}$ & CS$^{\uparrow}$ & CA$^{\uparrow}$ & FID$^{\downarrow}$ & CS$^{\uparrow}$ & CA$^{\uparrow}$ & FID$^{\downarrow}$ & \cellcolor{tablehighlight}CS$^{\downarrow}$ & \cellcolor{tablehighlight}CA$^{\downarrow}$ & \cellcolor{tablehighlight}FID$^{\uparrow}$ & CS$^{\uparrow}$ & CA$^{\uparrow}$ & FID$^{\downarrow}$ & CS$^{\uparrow}$ & CA$^{\uparrow}$ & FID$^{\downarrow}$ & CS$^{\uparrow}$ & CA$^{\uparrow}$ & FID$^{\downarrow}$ \\
    \midrule
    SPM & \cellcolor{tablehighlight}24.5 & \cellcolor{tablehighlight}85.8 & \cellcolor{tablehighlight}170 & 31.9 & \underline{99.6} & \textbf{40} & 31.0 & \underline{97.9} & \textbf{45} & 32.0 & \underline{99.6} & \textbf{22} & \cellcolor{tablehighlight}30.1 & \cellcolor{tablehighlight}92.5 & \cellcolor{tablehighlight}197 & \underline{30.7} & \textbf{99.6} & \textbf{22} & \underline{34.3} & 97.5 & \textbf{21} & 32.4 & \textbf{100.0} & \textbf{30} \\
    SAeUron & \cellcolor{tablehighlight}25.1 & \cellcolor{tablehighlight}77.5 & \cellcolor{tablehighlight}187 & 31.9 & \textbf{100.0} & 99 & 30.9 & 95.8 & 107 & 31.9 & 99.2 & 68 & \cellcolor{tablehighlight}31.2 & \cellcolor{tablehighlight}99.6 & \cellcolor{tablehighlight}101 & 30.3 & 97.1 & 72 & 34.1 & \underline{97.9} & 98 & \underline{32.5} & \underline{99.6} & 96 \\
    ESD-x & \cellcolor{tablehighlight}\underline{21.5} & \cellcolor{tablehighlight}55.8 & \cellcolor{tablehighlight}\underline{231} & 27.9 & 91.7 & 164 & 26.4 & 73.3 & 198 & 29.8 & 97.1 & 117 & \cellcolor{tablehighlight}23.4 & \cellcolor{tablehighlight}48.3 & \cellcolor{tablehighlight}238 & 25.2 & 82.1 & 158 & 30.9 & 93.3 & 131 & 30.1 & 96.2 & 151 \\
    ESD-u & \cellcolor{tablehighlight}22.3 & \cellcolor{tablehighlight}65.0 & \cellcolor{tablehighlight}220 & 25.4 & 84.6 & 221 & 25.4 & 69.6 & 240 & 27.1 & 81.7 & 184 & \cellcolor{tablehighlight}26.0 & \cellcolor{tablehighlight}77.1 & \cellcolor{tablehighlight}237 & 26.5 & 86.7 & 177 & 31.0 & 93.8 & 135 & 30.6 & 97.9 & 177 \\
    UCE & \cellcolor{tablehighlight}25.4 & \cellcolor{tablehighlight}77.1 & \cellcolor{tablehighlight}173 & \textbf{32.7} & \textbf{100.0} & 76 & \textbf{31.4} & \textbf{100.0} & \underline{83} & \underline{32.1} & \textbf{100.0} & 59 & \cellcolor{tablehighlight}25.2 & \cellcolor{tablehighlight}72.9 & \cellcolor{tablehighlight}203 & \textbf{31.5} & \underline{98.8} & 71 & \textbf{34.4} & \textbf{99.6} & \underline{75} & \textbf{32.6} & \textbf{100.0} & 83 \\
    MACE & \cellcolor{tablehighlight}22.4 & \cellcolor{tablehighlight}62.1 & \cellcolor{tablehighlight}193 & 28.8 & 90.4 & 131 & 28.9 & 89.6 & 116 & 31.3 & 98.3 & 76 & \cellcolor{tablehighlight}24.2 & \cellcolor{tablehighlight}49.6 & \cellcolor{tablehighlight}223 & 28.4 & 94.2 & 103 & 31.1 & 92.1 & 112 & 29.9 & 94.2 & 168 \\
    RECE & \cellcolor{tablehighlight}\textbf{20.3} & \cellcolor{tablehighlight}35.4 & \cellcolor{tablehighlight}\textbf{254} & 31.3 & \underline{99.6} & \underline{71} & 28.3 & 81.7 & 166 & 31.4 & 99.2 & 62 & \cellcolor{tablehighlight}\underline{21.3} & \cellcolor{tablehighlight}49.6 & \cellcolor{tablehighlight}288 & 30.1 & 96.7 & \underline{64} & 32.6 & 97.1 & 81 & 32.3 & \textbf{100.0} & \underline{62} \\
    CASteer & \cellcolor{tablehighlight}22.1 & \cellcolor{tablehighlight}\underline{30.4} & \cellcolor{tablehighlight}227 & 30.6 & 96.7 & 172 & 30.2 & 94.2 & 143 & 31.8 & \textbf{100.0} & 61 & \cellcolor{tablehighlight}\textbf{21.0} & \cellcolor{tablehighlight}\textbf{35.0} & \cellcolor{tablehighlight}\underline{314} & 27.1 & 95.4 & 216 & 29.2 & 93.8 & 190 & 30.9 & 97.5 & 141 \\
    \midrule
\rowcolor{ourshighlight}
    \textbf{\ours} & 21.6 & \textbf{12.1} & 213 & \underline{32.0} & \underline{99.6} & 93 & \underline{31.3} & 96.2 & 91 & \textbf{32.3} & \underline{99.6} & \underline{55} & 23.2 & \underline{39.6} & \textbf{318} & 29.4 & \underline{98.8} & 109 & 31.7 & 97.5 & 131 & 32.1 & \textbf{100.0} & 87 \\
\midrule
    \multicolumn{13}{c}{\textit{Erasing \textbf{Snoopy} and \textbf{Mickey}}} & \multicolumn{12}{|c}{\textit{Erasing \textbf{Picasso}}} \\
\midrule
    & \cellcolor{tablehighlight}CS$^{\downarrow}$ & \cellcolor{tablehighlight}CA$^{\downarrow}$ & \cellcolor{tablehighlight}FID$^{\uparrow}$ & \cellcolor{tablehighlight}CS$^{\downarrow}$ & \cellcolor{tablehighlight}CA$^{\downarrow}$ & \cellcolor{tablehighlight}FID$^{\uparrow}$ & CS$^{\uparrow}$ & CA$^{\uparrow}$ & FID$^{\downarrow}$ & CS$^{\uparrow}$ & CA$^{\uparrow}$ & FID$^{\downarrow}$ & CS$^{\uparrow}$ & CA$^{\uparrow}$ & FID$^{\downarrow}$ & \cellcolor{tablehighlight}CS$^{\downarrow}$ & \cellcolor{tablehighlight}CA$^{\downarrow}$ & \cellcolor{tablehighlight}FID$^{\uparrow}$ & CS$^{\uparrow}$ & CA$^{\uparrow}$ & FID$^{\downarrow}$ & CS$^{\uparrow}$ & CA$^{\uparrow}$ & FID$^{\downarrow}$ \\
    \midrule
    SPM & \cellcolor{tablehighlight}24.4 & \cellcolor{tablehighlight}77.9 & \cellcolor{tablehighlight}175 & \cellcolor{tablehighlight}24.5 & \cellcolor{tablehighlight}75.0 & \cellcolor{tablehighlight}229 & \underline{30.9} & \underline{97.5} & \textbf{59} & 31.8 & \textbf{99.6} & \textbf{39} & \underline{31.6} & \underline{99.6} & \textbf{36} & \cellcolor{tablehighlight}25.9 & \cellcolor{tablehighlight}79.6 & \cellcolor{tablehighlight}211 & 34.4 & 97.9 & \textbf{16} & 32.4 & \textbf{100.0} & \textbf{20} \\
    SAeUron & \cellcolor{tablehighlight}25.0 & \cellcolor{tablehighlight}74.6 & \cellcolor{tablehighlight}188 & \cellcolor{tablehighlight}29.7 & \cellcolor{tablehighlight}92.1 & \cellcolor{tablehighlight}144 & \underline{30.9} & 96.2 & 106 & \underline{31.9} & \underline{99.2} & 68 & 31.4 & \underline{99.6} & 106 & \cellcolor{tablehighlight}28.6 & \cellcolor{tablehighlight}95.0 & \cellcolor{tablehighlight}125 & 33.6 & 98.8 & 97 & \underline{32.5} & \underline{99.6} & 96 \\
    ESD-x & \cellcolor{tablehighlight}22.5 & \cellcolor{tablehighlight}67.9 & \cellcolor{tablehighlight}215 & \cellcolor{tablehighlight}22.8 & \cellcolor{tablehighlight}53.3 & \cellcolor{tablehighlight}255 & 25.9 & 72.5 & 204 & 29.4 & 92.1 & 145 & 30.9 & 97.5 & 172 & \cellcolor{tablehighlight}22.9 & \cellcolor{tablehighlight}56.2 & \cellcolor{tablehighlight}231 & 33.2 & 97.9 & 87 & 31.6 & 99.2 & 106 \\
    ESD-u & \cellcolor{tablehighlight}23.1 & \cellcolor{tablehighlight}70.0 & \cellcolor{tablehighlight}193 & \cellcolor{tablehighlight}23.6 & \cellcolor{tablehighlight}71.2 & \cellcolor{tablehighlight}236 & 26.5 & 75.0 & 213 & 28.7 & 94.6 & 174 & 27.6 & 94.6 & 203 & \cellcolor{tablehighlight}23.1 & \cellcolor{tablehighlight}62.5 & \cellcolor{tablehighlight}217 & 30.6 & 97.1 & 176 & 30.4 & \underline{99.6} & 155 \\
    UCE & \cellcolor{tablehighlight}24.3 & \cellcolor{tablehighlight}75.4 & \cellcolor{tablehighlight}185 & \cellcolor{tablehighlight}24.7 & \cellcolor{tablehighlight}57.9 & \cellcolor{tablehighlight}233 & \underline{30.9} & \textbf{97.9} & \underline{82} & 31.8 & \underline{99.2} & \underline{55} & \underline{31.6} & \textbf{100.0} & \underline{59} & \cellcolor{tablehighlight}25.1 & \cellcolor{tablehighlight}59.2 & \cellcolor{tablehighlight}148 & \textbf{34.8} & \underline{99.2} & \underline{53} & \textbf{32.6} & \textbf{100.0} & 60 \\
    MACE & \cellcolor{tablehighlight}22.4 & \cellcolor{tablehighlight}68.8 & \cellcolor{tablehighlight}198 & \cellcolor{tablehighlight}\underline{22.5} & \cellcolor{tablehighlight}\underline{37.5} & \cellcolor{tablehighlight}224 & 28.6 & 90.0 & 129 & 30.2 & 90.8 & 124 & 30.2 & 97.5 & 109 & \cellcolor{tablehighlight}23.6 & \cellcolor{tablehighlight}44.6 & \cellcolor{tablehighlight}190 & 32.5 & 96.2 & 89 & 31.4 & \textbf{100.0} & 122 \\
    RECE & \cellcolor{tablehighlight}\textbf{19.8} & \cellcolor{tablehighlight}38.3 & \cellcolor{tablehighlight}\textbf{250} & \cellcolor{tablehighlight}\textbf{19.6} & \cellcolor{tablehighlight}48.8 & \cellcolor{tablehighlight}\textbf{295} & 30.6 & 96.7 & 96 & 31.3 & 98.3 & 65 & \textbf{31.9} & \underline{99.6} & 62 & \cellcolor{tablehighlight}\textbf{20.5} & \cellcolor{tablehighlight}\textbf{33.3} & \cellcolor{tablehighlight}256 & 33.6 & 98.8 & 69 & 32.1 & \textbf{100.0} & 65 \\
    CASteer & \cellcolor{tablehighlight}22.1 & \cellcolor{tablehighlight}\underline{31.7} & \cellcolor{tablehighlight}\underline{226} & \cellcolor{tablehighlight}23.8 & \cellcolor{tablehighlight}66.2 & \cellcolor{tablehighlight}\underline{262} & 29.5 & 92.9 & 169 & 31.6 & \underline{99.2} & 81 & 26.0 & 92.5 & 294 & \cellcolor{tablehighlight}\underline{21.7} & \cellcolor{tablehighlight}58.8 & \cellcolor{tablehighlight}\textbf{280} & 32.2 & 97.9 & 143 & 31.9 & \textbf{100.0} & 98 \\
    \midrule
\rowcolor{ourshighlight}
    \textbf{\ours} & \underline{22.0} & \textbf{10.0} & 206 & 24.2 & \textbf{33.8} & 239 & \textbf{31.1} & 96.2 & 104 & \textbf{32.3} & \underline{99.2} & 62 & \underline{31.6} & \underline{99.6} & 71 & 23.5 & \underline{42.9} & \underline{263} & \underline{34.6} & \textbf{99.6} & 58 & \underline{32.5} & \textbf{100.0} & \underline{59} \\
\midrule
    \multicolumn{13}{c}{\textit{Erasing \textbf{Snoopy}, \textbf{Mickey} and \textbf{Spongebob}}} & \multicolumn{12}{|c}{\textit{Erasing \textbf{Rembrandt}}} \\
\midrule
    & \cellcolor{tablehighlight}CS$^{\downarrow}$ & \cellcolor{tablehighlight}CA$^{\downarrow}$ & \cellcolor{tablehighlight}FID$^{\uparrow}$ & \cellcolor{tablehighlight}CS$^{\downarrow}$ & \cellcolor{tablehighlight}CA$^{\downarrow}$ & \cellcolor{tablehighlight}FID$^{\uparrow}$ & \cellcolor{tablehighlight}CS$^{\downarrow}$ & \cellcolor{tablehighlight}CA$^{\downarrow}$ & \cellcolor{tablehighlight}FID$^{\uparrow}$ & CS$^{\uparrow}$ & CA$^{\uparrow}$ & FID$^{\downarrow}$ & CS$^{\uparrow}$ & CA$^{\uparrow}$ & FID$^{\downarrow}$ & CS$^{\uparrow}$ & CA$^{\uparrow}$ & FID$^{\downarrow}$ & \cellcolor{tablehighlight}CS$^{\downarrow}$ & \cellcolor{tablehighlight}CA$^{\downarrow}$ & \cellcolor{tablehighlight}FID$^{\uparrow}$ & CS$^{\uparrow}$ & CA$^{\uparrow}$ & FID$^{\downarrow}$ \\
    \midrule
    SPM & \cellcolor{tablehighlight}24.6 & \cellcolor{tablehighlight}83.8 & \cellcolor{tablehighlight}173 & \cellcolor{tablehighlight}24.7 & \cellcolor{tablehighlight}75.0 & \cellcolor{tablehighlight}226 & \cellcolor{tablehighlight}25.4 & \cellcolor{tablehighlight}67.9 & \cellcolor{tablehighlight}224 & 31.8 & \underline{99.6} & \textbf{42} & 31.7 & \underline{99.6} & \textbf{34} & \underline{30.6} & \textbf{99.2} & \textbf{21} & \cellcolor{tablehighlight}26.7 & \cellcolor{tablehighlight}70.4 & \cellcolor{tablehighlight}173 & \textbf{32.4} & \textbf{100.0} & \textbf{22} \\
    SAeUron & \cellcolor{tablehighlight}25.1 & \cellcolor{tablehighlight}78.3 & \cellcolor{tablehighlight}188 & \cellcolor{tablehighlight}28.9 & \cellcolor{tablehighlight}86.2 & \cellcolor{tablehighlight}155 & \cellcolor{tablehighlight}29.0 & \cellcolor{tablehighlight}84.6 & \cellcolor{tablehighlight}135 & 31.8 & 99.2 & 71 & 31.4 & \underline{99.6} & 99 & 30.4 & 96.7 & 72 & \cellcolor{tablehighlight}33.4 & \cellcolor{tablehighlight}97.5 & \cellcolor{tablehighlight}99 & \textbf{32.4} & \underline{99.6} & 98 \\
    ESD-x & \cellcolor{tablehighlight}23.2 & \cellcolor{tablehighlight}74.6 & \cellcolor{tablehighlight}195 & \cellcolor{tablehighlight}22.7 & \cellcolor{tablehighlight}60.8 & \cellcolor{tablehighlight}235 & \cellcolor{tablehighlight}24.7 & \cellcolor{tablehighlight}64.2 & \cellcolor{tablehighlight}240 & 28.9 & 92.5 & 145 & 28.2 & 86.7 & 175 & 23.8 & 72.5 & 189 & \cellcolor{tablehighlight}21.1 & \cellcolor{tablehighlight}12.1 & \cellcolor{tablehighlight}241 & 30.5 & 96.7 & 165 \\
    ESD-u & \cellcolor{tablehighlight}24.0 & \cellcolor{tablehighlight}79.6 & \cellcolor{tablehighlight}194 & \cellcolor{tablehighlight}23.1 & \cellcolor{tablehighlight}70.4 & \cellcolor{tablehighlight}229 & \cellcolor{tablehighlight}24.3 & \cellcolor{tablehighlight}60.4 & \cellcolor{tablehighlight}237 & 27.3 & 85.8 & 172 & 29.2 & 97.1 & 183 & 26.2 & 91.7 & 153 & \cellcolor{tablehighlight}24.1 & \cellcolor{tablehighlight}51.7 & \cellcolor{tablehighlight}208 & \textbf{32.4} & 97.9 & 184 \\
    UCE & \cellcolor{tablehighlight}24.4 & \cellcolor{tablehighlight}74.2 & \cellcolor{tablehighlight}177 & \cellcolor{tablehighlight}24.9 & \cellcolor{tablehighlight}53.3 & \cellcolor{tablehighlight}223 & \cellcolor{tablehighlight}23.5 & \cellcolor{tablehighlight}43.8 & \cellcolor{tablehighlight}221 & \underline{31.9} & \textbf{100.0} & 63 & \underline{31.8} & \underline{99.6} & \underline{63} & \textbf{30.9} & \textbf{99.2} & \underline{43} & \cellcolor{tablehighlight}25.0 & \cellcolor{tablehighlight}41.7 & \cellcolor{tablehighlight}179 & \textbf{32.4} & \textbf{100.0} & \underline{54} \\
    MACE & \cellcolor{tablehighlight}22.3 & \cellcolor{tablehighlight}63.8 & \cellcolor{tablehighlight}201 & \cellcolor{tablehighlight}\underline{22.2} & \cellcolor{tablehighlight}\textbf{35.0} & \cellcolor{tablehighlight}225 & \cellcolor{tablehighlight}21.7 & \cellcolor{tablehighlight}31.7 & \cellcolor{tablehighlight}203 & 26.6 & 70.4 & 176 & 28.4 & 94.6 & 139 & 27.4 & 92.1 & 120 & \cellcolor{tablehighlight}23.4 & \cellcolor{tablehighlight}32.9 & \cellcolor{tablehighlight}224 & 31.5 & 97.9 & 87 \\
    RECE & \cellcolor{tablehighlight}\textbf{20.2} & \cellcolor{tablehighlight}37.5 & \cellcolor{tablehighlight}\textbf{254} & \cellcolor{tablehighlight}\textbf{20.1} & \cellcolor{tablehighlight}50.0 & \cellcolor{tablehighlight}\textbf{307} & \cellcolor{tablehighlight}\textbf{18.2} & \cellcolor{tablehighlight}\textbf{11.2} & \cellcolor{tablehighlight}\underline{321} & 31.4 & 97.5 & 71 & \textbf{31.9} & \textbf{100.0} & 77 & 29.9 & \underline{98.3} & 85 & \cellcolor{tablehighlight}21.3 & \cellcolor{tablehighlight}20.8 & \cellcolor{tablehighlight}217 & \underline{32.2} & \textbf{100.0} & 89 \\
    CASteer & \cellcolor{tablehighlight}22.5 & \cellcolor{tablehighlight}\underline{35.4} & \cellcolor{tablehighlight}\underline{231} & \cellcolor{tablehighlight}23.8 & \cellcolor{tablehighlight}69.6 & \cellcolor{tablehighlight}\underline{277} & \cellcolor{tablehighlight}22.9 & \cellcolor{tablehighlight}32.5 & \cellcolor{tablehighlight}\textbf{329} & 31.4 & 98.3 & 94 & 24.1 & 80.4 & 289 & 25.2 & 88.3 & 187 & \cellcolor{tablehighlight}\underline{18.6} & \cellcolor{tablehighlight}\underline{5.0} & \cellcolor{tablehighlight}\underline{336} & 31.1 & \textbf{100.0} & 118 \\
    \midrule
\rowcolor{ourshighlight}
    \textbf{\ours} & \underline{21.8} & \textbf{15.4} & 209 & 24.3 & \underline{49.2} & 231 & \underline{21.5} & \underline{13.3} & 304 & \textbf{32.2} & 99.2 & \underline{61} & 29.4 & 97.5 & 158 & 28.5 & 95.8 & 112 & \textbf{18.0} & \textbf{1.7} & \textbf{350} & 31.6 & \textbf{100.0} & 102 \\
    \bottomrule
\end{tabular}
\end{table}

\begin{table}[t]
\centering
\caption{\textbf{Erasure of instances and artistic styles on SD-v2.1.} Every method is
re-run from its official code on this backbone; shaded columns mark the erased concepts; blue rows mark \ours\ results. Bold marks
the best value and underline the second best. \ours\ uses $\beta=3$.}
\label{tab:xmodel_sd21}
\fontsize{6.6}{8.2}\selectfont
\setlength{\tabcolsep}{0.6pt}
\setlength{\heavyrulewidth}{0.6pt}
\setlength{\lightrulewidth}{0.3pt}
\begin{tabular}{@{}c|ccc|ccc|ccc|ccc|ccc|ccc|ccc|ccc@{}}
    \toprule
    Method & \multicolumn{3}{c|}{Snoopy} & \multicolumn{3}{c|}{Mickey} & \multicolumn{3}{c|}{Spongebob} & \multicolumn{3}{c|}{Pikachu} & \multicolumn{3}{c|}{Van Gogh} & \multicolumn{3}{c|}{Picasso} & \multicolumn{3}{c|}{Rembrandt} & \multicolumn{3}{c}{Hokusai} \\
    & CS & CA & FID & CS & CA & FID & CS & CA & FID & CS & CA & FID & CS & CA & FID & CS & CA & FID & CS & CA & FID & CS & CA & FID \\
    \midrule
    SD & 32.4 & 99.6 & 0 & 32.2 & 99.2 & 0 & 29.5 & 98.3 & 0 & 31.7 & 100.0 & 0 & 32.0 & 100.0 & 0 & 32.2 & 97.5 & 0 & 33.1 & 96.7 & 0 & 32.2 & 100.0 & 0 \\
\midrule
    \multicolumn{13}{c}{\textit{Erasing \textbf{Snoopy}}} & \multicolumn{12}{|c}{\textit{Erasing \textbf{Van Gogh}}} \\
\midrule
    & \cellcolor{tablehighlight}CS$^{\downarrow}$ & \cellcolor{tablehighlight}CA$^{\downarrow}$ & \cellcolor{tablehighlight}FID$^{\uparrow}$ & CS$^{\uparrow}$ & CA$^{\uparrow}$ & FID$^{\downarrow}$ & CS$^{\uparrow}$ & CA$^{\uparrow}$ & FID$^{\downarrow}$ & CS$^{\uparrow}$ & CA$^{\uparrow}$ & FID$^{\downarrow}$ & \cellcolor{tablehighlight}CS$^{\downarrow}$ & \cellcolor{tablehighlight}CA$^{\downarrow}$ & \cellcolor{tablehighlight}FID$^{\uparrow}$ & CS$^{\uparrow}$ & CA$^{\uparrow}$ & FID$^{\downarrow}$ & CS$^{\uparrow}$ & CA$^{\uparrow}$ & FID$^{\downarrow}$ & CS$^{\uparrow}$ & CA$^{\uparrow}$ & FID$^{\downarrow}$ \\
    \midrule
    SPM & \cellcolor{tablehighlight}24.7 & \cellcolor{tablehighlight}69.6 & \cellcolor{tablehighlight}204 & 31.9 & \underline{98.8} & \underline{65} & 29.2 & \textbf{97.9} & \underline{80} & 31.7 & \textbf{100.0} & \textbf{30} & \cellcolor{tablehighlight}26.8 & \cellcolor{tablehighlight}75.0 & \cellcolor{tablehighlight}248 & \underline{31.9} & 97.1 & \underline{42} & \underline{32.7} & 97.5 & \underline{28} & \textbf{32.2} & \textbf{100.0} & \textbf{50} \\
    SAeUron & \cellcolor{tablehighlight}31.8 & \cellcolor{tablehighlight}97.5 & \cellcolor{tablehighlight}95 & 32.2 & 97.5 & 99 & 29.3 & 91.2 & 127 & 31.9 & 99.2 & 75 & \cellcolor{tablehighlight}31.5 & \cellcolor{tablehighlight}99.6 & \cellcolor{tablehighlight}82 & 31.2 & \textbf{98.3} & 73 & 32.4 & 95.8 & 61 & 32.0 & \underline{99.6} & 102 \\
    ESD-x & \cellcolor{tablehighlight}\underline{23.0} & \cellcolor{tablehighlight}63.8 & \cellcolor{tablehighlight}220 & 27.6 & 93.3 & 175 & 26.5 & 74.2 & 165 & 31.3 & 99.2 & 90 & \cellcolor{tablehighlight}25.7 & \cellcolor{tablehighlight}82.1 & \cellcolor{tablehighlight}245 & 29.0 & 94.2 & 107 & 31.4 & 95.8 & 123 & 30.5 & 97.5 & 151 \\
    ESD-u & \cellcolor{tablehighlight}25.4 & \cellcolor{tablehighlight}81.7 & \cellcolor{tablehighlight}190 & 28.6 & 93.8 & 162 & 26.7 & 77.1 & 160 & 31.0 & \underline{99.6} & 102 & \cellcolor{tablehighlight}26.7 & \cellcolor{tablehighlight}90.8 & \cellcolor{tablehighlight}213 & 28.8 & 97.1 & 132 & 30.9 & 95.4 & 150 & 30.9 & 99.2 & 168 \\
    UCE & \cellcolor{tablehighlight}27.6 & \cellcolor{tablehighlight}90.4 & \cellcolor{tablehighlight}128 & \underline{32.3} & \textbf{99.2} & \textbf{42} & 29.4 & \textbf{97.9} & \textbf{63} & 31.7 & \underline{99.6} & \underline{31} & \cellcolor{tablehighlight}31.5 & \cellcolor{tablehighlight}98.3 & \cellcolor{tablehighlight}191 & \textbf{32.2} & 97.1 & \textbf{30} & \textbf{33.1} & 97.5 & \textbf{19} & \underline{32.1} & \textbf{100.0} & \underline{51} \\
    MACE & \cellcolor{tablehighlight}24.8 & \cellcolor{tablehighlight}82.5 & \cellcolor{tablehighlight}194 & 30.4 & \underline{98.8} & 114 & 28.4 & 94.2 & 115 & 31.1 & \underline{99.6} & 68 & \cellcolor{tablehighlight}28.5 & \cellcolor{tablehighlight}90.8 & \cellcolor{tablehighlight}122 & 30.9 & 96.7 & 74 & 32.1 & \underline{97.9} & 66 & 31.2 & 99.2 & 104 \\
    RECE & \cellcolor{tablehighlight}\textbf{19.0} & \cellcolor{tablehighlight}\underline{30.4} & \cellcolor{tablehighlight}\textbf{273} & 28.9 & 97.1 & 124 & 26.9 & 78.3 & 142 & 30.0 & 97.9 & 102 & \cellcolor{tablehighlight}\textbf{18.9} & \cellcolor{tablehighlight}\underline{22.1} & \cellcolor{tablehighlight}327 & 31.1 & \underline{97.5} & 67 & 31.9 & \textbf{98.3} & 69 & 32.0 & \textbf{100.0} & 87 \\
    CASteer & \cellcolor{tablehighlight}23.7 & \cellcolor{tablehighlight}44.6 & \cellcolor{tablehighlight}223 & 31.5 & 97.9 & 128 & \underline{29.5} & 94.2 & 118 & \underline{32.3} & \textbf{100.0} & 54 & \cellcolor{tablehighlight}\underline{21.4} & \cellcolor{tablehighlight}34.2 & \cellcolor{tablehighlight}\textbf{359} & 30.1 & 97.1 & 136 & 29.9 & 96.2 & 138 & 28.6 & 96.7 & 231 \\
    \midrule
\rowcolor{ourshighlight}
    \textbf{\ours} & \underline{23.0} & \textbf{17.5} & \underline{233} & \textbf{32.5} & \textbf{99.2} & 113 & \textbf{30.1} & \underline{97.1} & 119 & \textbf{32.7} & \textbf{100.0} & 69 & 21.8 & \textbf{18.3} & \underline{332} & 30.6 & 96.7 & 140 & 30.7 & 95.8 & 157 & 28.0 & 91.7 & 243 \\
\midrule
    \multicolumn{13}{c}{\textit{Erasing \textbf{Snoopy} and \textbf{Mickey}}} & \multicolumn{12}{|c}{\textit{Erasing \textbf{Picasso}}} \\
\midrule
    & \cellcolor{tablehighlight}CS$^{\downarrow}$ & \cellcolor{tablehighlight}CA$^{\downarrow}$ & \cellcolor{tablehighlight}FID$^{\uparrow}$ & \cellcolor{tablehighlight}CS$^{\downarrow}$ & \cellcolor{tablehighlight}CA$^{\downarrow}$ & \cellcolor{tablehighlight}FID$^{\uparrow}$ & CS$^{\uparrow}$ & CA$^{\uparrow}$ & FID$^{\downarrow}$ & CS$^{\uparrow}$ & CA$^{\uparrow}$ & FID$^{\downarrow}$ & CS$^{\uparrow}$ & CA$^{\uparrow}$ & FID$^{\downarrow}$ & \cellcolor{tablehighlight}CS$^{\downarrow}$ & \cellcolor{tablehighlight}CA$^{\downarrow}$ & \cellcolor{tablehighlight}FID$^{\uparrow}$ & CS$^{\uparrow}$ & CA$^{\uparrow}$ & FID$^{\downarrow}$ & CS$^{\uparrow}$ & CA$^{\uparrow}$ & FID$^{\downarrow}$ \\
    \midrule
    SPM & \cellcolor{tablehighlight}25.7 & \cellcolor{tablehighlight}75.0 & \cellcolor{tablehighlight}187 & \cellcolor{tablehighlight}28.4 & \cellcolor{tablehighlight}82.1 & \cellcolor{tablehighlight}162 & 29.2 & \textbf{97.5} & \textbf{72} & 31.7 & \textbf{100.0} & \textbf{32} & \underline{31.9} & \textbf{100.0} & \textbf{32} & \cellcolor{tablehighlight}30.8 & \cellcolor{tablehighlight}95.0 & \cellcolor{tablehighlight}70 & \underline{33.1} & 97.5 & \textbf{18} & \textbf{32.3} & \textbf{100.0} & \textbf{38} \\
    SAeUron & \cellcolor{tablehighlight}31.8 & \cellcolor{tablehighlight}97.5 & \cellcolor{tablehighlight}96 & \cellcolor{tablehighlight}32.2 & \cellcolor{tablehighlight}97.9 & \cellcolor{tablehighlight}99 & 29.2 & 90.8 & 129 & 32.0 & 99.2 & 74 & 31.5 & \underline{99.6} & 82 & \cellcolor{tablehighlight}31.3 & \cellcolor{tablehighlight}97.9 & \cellcolor{tablehighlight}72 & 32.4 & 96.2 & 61 & 32.0 & \underline{99.6} & 103 \\
    ESD-x & \cellcolor{tablehighlight}25.1 & \cellcolor{tablehighlight}82.5 & \cellcolor{tablehighlight}197 & \cellcolor{tablehighlight}25.3 & \cellcolor{tablehighlight}85.4 & \cellcolor{tablehighlight}218 & 27.8 & 85.4 & 146 & 31.4 & 99.2 & 89 & 30.2 & 99.2 & 112 & \cellcolor{tablehighlight}\underline{23.7} & \cellcolor{tablehighlight}79.2 & \cellcolor{tablehighlight}236 & 32.4 & 95.0 & 115 & 31.3 & 98.8 & 105 \\
    ESD-u & \cellcolor{tablehighlight}25.2 & \cellcolor{tablehighlight}82.5 & \cellcolor{tablehighlight}200 & \cellcolor{tablehighlight}24.7 & \cellcolor{tablehighlight}80.0 & \cellcolor{tablehighlight}\underline{244} & 26.6 & 65.0 & 176 & 30.4 & 98.8 & 131 & 29.8 & 98.8 & 135 & \cellcolor{tablehighlight}24.3 & \cellcolor{tablehighlight}71.7 & \cellcolor{tablehighlight}200 & 31.1 & 94.2 & 129 & 31.2 & 98.3 & 102 \\
    UCE & \cellcolor{tablehighlight}27.4 & \cellcolor{tablehighlight}91.7 & \cellcolor{tablehighlight}137 & \cellcolor{tablehighlight}28.0 & \cellcolor{tablehighlight}86.2 & \cellcolor{tablehighlight}153 & \underline{29.4} & \underline{97.1} & \underline{78} & 31.8 & \textbf{100.0} & \underline{39} & \textbf{32.1} & \textbf{100.0} & \underline{36} & \cellcolor{tablehighlight}30.7 & \cellcolor{tablehighlight}94.6 & \cellcolor{tablehighlight}66 & \underline{33.1} & 97.5 & \underline{21} & \underline{32.2} & \textbf{100.0} & \underline{52} \\
    MACE & \cellcolor{tablehighlight}23.9 & \cellcolor{tablehighlight}74.6 & \cellcolor{tablehighlight}195 & \cellcolor{tablehighlight}27.2 & \cellcolor{tablehighlight}86.7 & \cellcolor{tablehighlight}177 & 28.1 & 92.5 & 145 & 30.9 & \underline{99.6} & 100 & 30.4 & \underline{99.6} & 78 & \cellcolor{tablehighlight}27.5 & \cellcolor{tablehighlight}78.3 & \cellcolor{tablehighlight}127 & 32.7 & 97.9 & 48 & 31.8 & \underline{99.6} & 75 \\
    RECE & \cellcolor{tablehighlight}\textbf{19.0} & \cellcolor{tablehighlight}\underline{35.8} & \cellcolor{tablehighlight}\textbf{286} & \cellcolor{tablehighlight}\textbf{19.6} & \cellcolor{tablehighlight}\underline{72.5} & \cellcolor{tablehighlight}\textbf{320} & 26.0 & 70.4 & 157 & 30.0 & 98.3 & 86 & 30.3 & \textbf{100.0} & 75 & \cellcolor{tablehighlight}\textbf{19.4} & \cellcolor{tablehighlight}\textbf{30.8} & \cellcolor{tablehighlight}\textbf{303} & 33.0 & \textbf{99.2} & 55 & 31.9 & \textbf{100.0} & 86 \\
    CASteer & \cellcolor{tablehighlight}23.7 & \cellcolor{tablehighlight}43.3 & \cellcolor{tablehighlight}227 & \cellcolor{tablehighlight}27.6 & \cellcolor{tablehighlight}80.8 & \cellcolor{tablehighlight}207 & \underline{29.4} & 92.9 & 133 & \underline{32.2} & \textbf{100.0} & 62 & 27.1 & 97.5 & 288 & \cellcolor{tablehighlight}24.7 & \cellcolor{tablehighlight}66.2 & \cellcolor{tablehighlight}253 & 32.0 & 97.9 & 118 & 31.7 & \textbf{100.0} & 86 \\
    \midrule
\rowcolor{ourshighlight}
    \textbf{\ours} & \underline{22.7} & \textbf{15.4} & \underline{233} & \underline{24.6} & \textbf{40.8} & 241 & \textbf{30.1} & 95.4 & 111 & \textbf{32.9} & \textbf{100.0} & 69 & 30.2 & 99.2 & 158 & 24.2 & \underline{45.4} & \underline{296} & \textbf{33.8} & \underline{98.8} & 64 & \textbf{32.3} & \textbf{100.0} & 86 \\
\midrule
    \multicolumn{13}{c}{\textit{Erasing \textbf{Snoopy}, \textbf{Mickey} and \textbf{Spongebob}}} & \multicolumn{12}{|c}{\textit{Erasing \textbf{Rembrandt}}} \\
\midrule
    & \cellcolor{tablehighlight}CS$^{\downarrow}$ & \cellcolor{tablehighlight}CA$^{\downarrow}$ & \cellcolor{tablehighlight}FID$^{\uparrow}$ & \cellcolor{tablehighlight}CS$^{\downarrow}$ & \cellcolor{tablehighlight}CA$^{\downarrow}$ & \cellcolor{tablehighlight}FID$^{\uparrow}$ & \cellcolor{tablehighlight}CS$^{\downarrow}$ & \cellcolor{tablehighlight}CA$^{\downarrow}$ & \cellcolor{tablehighlight}FID$^{\uparrow}$ & CS$^{\uparrow}$ & CA$^{\uparrow}$ & FID$^{\downarrow}$ & CS$^{\uparrow}$ & CA$^{\uparrow}$ & FID$^{\downarrow}$ & CS$^{\uparrow}$ & CA$^{\uparrow}$ & FID$^{\downarrow}$ & \cellcolor{tablehighlight}CS$^{\downarrow}$ & \cellcolor{tablehighlight}CA$^{\downarrow}$ & \cellcolor{tablehighlight}FID$^{\uparrow}$ & CS$^{\uparrow}$ & CA$^{\uparrow}$ & FID$^{\downarrow}$ \\
    \midrule
    SPM & \cellcolor{tablehighlight}25.0 & \cellcolor{tablehighlight}76.2 & \cellcolor{tablehighlight}192 & \cellcolor{tablehighlight}29.2 & \cellcolor{tablehighlight}89.6 & \cellcolor{tablehighlight}143 & \cellcolor{tablehighlight}27.6 & \cellcolor{tablehighlight}87.1 & \cellcolor{tablehighlight}143 & 31.6 & \textbf{100.0} & \textbf{36} & \underline{31.7} & \textbf{100.0} & \textbf{34} & \textbf{32.2} & 97.5 & \textbf{31} & \cellcolor{tablehighlight}31.0 & \cellcolor{tablehighlight}92.9 & \cellcolor{tablehighlight}103 & \textbf{32.3} & \textbf{100.0} & \textbf{41} \\
    SAeUron & \cellcolor{tablehighlight}31.9 & \cellcolor{tablehighlight}96.2 & \cellcolor{tablehighlight}95 & \cellcolor{tablehighlight}32.2 & \cellcolor{tablehighlight}97.9 & \cellcolor{tablehighlight}99 & \cellcolor{tablehighlight}29.3 & \cellcolor{tablehighlight}91.2 & \cellcolor{tablehighlight}129 & \underline{32.0} & \underline{99.2} & 74 & 31.6 & \underline{99.6} & 82 & 31.2 & \underline{98.8} & 72 & \cellcolor{tablehighlight}32.4 & \cellcolor{tablehighlight}96.7 & \cellcolor{tablehighlight}61 & 32.0 & 99.2 & 104 \\
    ESD-x & \cellcolor{tablehighlight}25.5 & \cellcolor{tablehighlight}85.4 & \cellcolor{tablehighlight}184 & \cellcolor{tablehighlight}25.4 & \cellcolor{tablehighlight}89.6 & \cellcolor{tablehighlight}212 & \cellcolor{tablehighlight}25.3 & \cellcolor{tablehighlight}61.2 & \cellcolor{tablehighlight}192 & 30.9 & \underline{99.2} & 107 & 30.1 & 99.2 & 84 & 29.5 & 96.2 & 96 & \cellcolor{tablehighlight}26.0 & \cellcolor{tablehighlight}62.5 & \cellcolor{tablehighlight}218 & 31.9 & \textbf{100.0} & 95 \\
    ESD-u & \cellcolor{tablehighlight}25.9 & \cellcolor{tablehighlight}83.8 & \cellcolor{tablehighlight}186 & \cellcolor{tablehighlight}25.6 & \cellcolor{tablehighlight}82.5 & \cellcolor{tablehighlight}220 & \cellcolor{tablehighlight}25.4 & \cellcolor{tablehighlight}64.6 & \cellcolor{tablehighlight}199 & 29.4 & 95.8 & 158 & 28.7 & 98.8 & 162 & 26.0 & 90.0 & 195 & \cellcolor{tablehighlight}24.0 & \cellcolor{tablehighlight}52.5 & \cellcolor{tablehighlight}257 & 31.6 & \underline{99.6} & 128 \\
    UCE & \cellcolor{tablehighlight}27.3 & \cellcolor{tablehighlight}92.5 & \cellcolor{tablehighlight}142 & \cellcolor{tablehighlight}27.7 & \cellcolor{tablehighlight}87.1 & \cellcolor{tablehighlight}151 & \cellcolor{tablehighlight}25.1 & \cellcolor{tablehighlight}40.0 & \cellcolor{tablehighlight}154 & 31.8 & \textbf{100.0} & \underline{45} & \textbf{31.9} & \textbf{100.0} & \underline{44} & \textbf{32.2} & 95.8 & \underline{41} & \cellcolor{tablehighlight}31.5 & \cellcolor{tablehighlight}96.2 & \cellcolor{tablehighlight}85 & \underline{32.1} & \textbf{100.0} & \underline{56} \\
    MACE & \cellcolor{tablehighlight}22.9 & \cellcolor{tablehighlight}70.8 & \cellcolor{tablehighlight}209 & \cellcolor{tablehighlight}26.7 & \cellcolor{tablehighlight}83.8 & \cellcolor{tablehighlight}185 & \cellcolor{tablehighlight}\underline{22.2} & \cellcolor{tablehighlight}45.8 & \cellcolor{tablehighlight}212 & 28.5 & 93.3 & 169 & 29.7 & \textbf{100.0} & 84 & 31.6 & 98.3 & 63 & \cellcolor{tablehighlight}25.9 & \cellcolor{tablehighlight}55.4 & \cellcolor{tablehighlight}125 & 31.8 & \textbf{100.0} & 80 \\
    RECE & \cellcolor{tablehighlight}\textbf{19.4} & \cellcolor{tablehighlight}\underline{24.6} & \cellcolor{tablehighlight}\textbf{282} & \cellcolor{tablehighlight}\textbf{18.7} & \cellcolor{tablehighlight}\textbf{49.6} & \cellcolor{tablehighlight}\textbf{345} & \cellcolor{tablehighlight}\textbf{17.9} & \cellcolor{tablehighlight}\textbf{2.5} & \cellcolor{tablehighlight}\textbf{304} & 31.6 & \textbf{100.0} & 61 & 30.1 & \textbf{100.0} & 75 & \underline{31.8} & \textbf{99.2} & 61 & \cellcolor{tablehighlight}20.4 & \cellcolor{tablehighlight}15.4 & \cellcolor{tablehighlight}282 & 32.0 & \textbf{100.0} & 82 \\
    CASteer & \cellcolor{tablehighlight}23.7 & \cellcolor{tablehighlight}46.7 & \cellcolor{tablehighlight}226 & \cellcolor{tablehighlight}27.6 & \cellcolor{tablehighlight}79.2 & \cellcolor{tablehighlight}210 & \cellcolor{tablehighlight}24.6 & \cellcolor{tablehighlight}60.4 & \cellcolor{tablehighlight}238 & 31.8 & 98.3 & 85 & 25.9 & 95.8 & 287 & 28.4 & 89.2 & 141 & \cellcolor{tablehighlight}\underline{19.2} & \cellcolor{tablehighlight}\underline{11.2} & \cellcolor{tablehighlight}\underline{340} & 30.9 & 97.9 & 151 \\
    \midrule
\rowcolor{ourshighlight}
    \textbf{\ours} & \underline{22.8} & \textbf{23.8} & \underline{239} & \underline{24.9} & \underline{53.3} & \underline{249} & 23.6 & \underline{11.2} & \underline{255} & \textbf{32.3} & 98.8 & 105 & 26.8 & 96.7 & 231 & 27.4 & 90.4 & 152 & \textbf{18.2} & \textbf{5.8} & \textbf{361} & 31.3 & 98.3 & 115 \\
    \bottomrule
\end{tabular}
\end{table}

\begin{table}[t]
\centering
\caption{\textbf{Erasure of instances and artistic styles on SDXL.} Every method is
re-run from its official code on this backbone; shaded columns mark the erased concepts; blue rows mark \ours\ results. Bold marks
the best value and underline the second best. \ours\ uses $\beta=3$.}
\label{tab:xmodel_sdxl}
\fontsize{6.6}{8.2}\selectfont
\setlength{\tabcolsep}{0.6pt}
\setlength{\heavyrulewidth}{0.6pt}
\setlength{\lightrulewidth}{0.3pt}
\begin{tabular}{@{}c|ccc|ccc|ccc|ccc|ccc|ccc|ccc|ccc@{}}
    \toprule
    Method & \multicolumn{3}{c|}{Snoopy} & \multicolumn{3}{c|}{Mickey} & \multicolumn{3}{c|}{Spongebob} & \multicolumn{3}{c|}{Pikachu} & \multicolumn{3}{c|}{Van Gogh} & \multicolumn{3}{c|}{Picasso} & \multicolumn{3}{c|}{Rembrandt} & \multicolumn{3}{c}{Hokusai} \\
    & CS & CA & FID & CS & CA & FID & CS & CA & FID & CS & CA & FID & CS & CA & FID & CS & CA & FID & CS & CA & FID & CS & CA & FID \\
    \midrule
    SD & 32.4 & 100.0 & 0 & 32.0 & 99.2 & 0 & 30.6 & 93.3 & 0 & 31.2 & 97.5 & 0 & 28.7 & 95.8 & 0 & 29.4 & 98.8 & 0 & 29.9 & 95.0 & 0 & 31.7 & 100.0 & 0 \\
\midrule
    \multicolumn{13}{c}{\textit{Erasing \textbf{Snoopy}}} & \multicolumn{12}{|c}{\textit{Erasing \textbf{Van Gogh}}} \\
\midrule
    & \cellcolor{tablehighlight}CS$^{\downarrow}$ & \cellcolor{tablehighlight}CA$^{\downarrow}$ & \cellcolor{tablehighlight}FID$^{\uparrow}$ & CS$^{\uparrow}$ & CA$^{\uparrow}$ & FID$^{\downarrow}$ & CS$^{\uparrow}$ & CA$^{\uparrow}$ & FID$^{\downarrow}$ & CS$^{\uparrow}$ & CA$^{\uparrow}$ & FID$^{\downarrow}$ & \cellcolor{tablehighlight}CS$^{\downarrow}$ & \cellcolor{tablehighlight}CA$^{\downarrow}$ & \cellcolor{tablehighlight}FID$^{\uparrow}$ & CS$^{\uparrow}$ & CA$^{\uparrow}$ & FID$^{\downarrow}$ & CS$^{\uparrow}$ & CA$^{\uparrow}$ & FID$^{\downarrow}$ & CS$^{\uparrow}$ & CA$^{\uparrow}$ & FID$^{\downarrow}$ \\
    \midrule
    ESD-x & \cellcolor{tablehighlight}23.6 & \cellcolor{tablehighlight}40.8 & \cellcolor{tablehighlight}\underline{269} & 26.8 & 71.2 & 229 & 26.2 & \textbf{92.9} & 316 & 28.7 & 70.0 & 265 & \cellcolor{tablehighlight}25.9 & \cellcolor{tablehighlight}93.3 & \cellcolor{tablehighlight}\textbf{288} & 26.3 & 87.5 & 139 & 27.1 & 85.0 & 125 & 28.2 & \textbf{100.0} & 229 \\
    ESD-u & \cellcolor{tablehighlight}31.4 & \cellcolor{tablehighlight}97.9 & \cellcolor{tablehighlight}65 & \underline{31.7} & 98.3 & \textbf{70} & \underline{30.2} & 90.8 & \underline{77} & 31.1 & \textbf{98.8} & \underline{43} & \cellcolor{tablehighlight}27.9 & \cellcolor{tablehighlight}96.7 & \cellcolor{tablehighlight}86 & \underline{28.4} & 96.7 & \underline{66} & 29.3 & \underline{94.6} & \underline{46} & \underline{31.6} & \textbf{100.0} & \underline{62} \\
    UCE & \cellcolor{tablehighlight}\underline{23.1} & \cellcolor{tablehighlight}\textbf{1.7} & \cellcolor{tablehighlight}\textbf{324} & \textbf{32.0} & \textbf{99.6} & \underline{88} & 27.5 & 76.2 & 158 & 30.3 & 96.7 & 106 & \cellcolor{tablehighlight}23.4 & \cellcolor{tablehighlight}77.9 & \cellcolor{tablehighlight}140 & \textbf{29.4} & \textbf{98.3} & \textbf{20} & \textbf{29.9} & \underline{94.6} & \textbf{15} & \textbf{31.7} & \textbf{100.0} & \textbf{21} \\
    CASteer & \cellcolor{tablehighlight}\underline{23.1} & \cellcolor{tablehighlight}45.0 & \cellcolor{tablehighlight}228 & 30.4 & 92.9 & 151 & \underline{30.2} & 90.8 & 108 & \textbf{31.3} & \textbf{98.8} & 52 & \cellcolor{tablehighlight}\underline{20.2} & \cellcolor{tablehighlight}\underline{27.9} & \cellcolor{tablehighlight}198 & 26.7 & \underline{97.1} & 122 & 28.9 & \textbf{95.0} & 92 & 30.8 & \textbf{100.0} & 98 \\
    \midrule
\rowcolor{ourshighlight}
    \textbf{\ours} & \textbf{22.4} & \underline{18.3} & 235 & \underline{31.7} & \underline{99.2} & \underline{88} & \textbf{30.6} & \underline{92.1} & \textbf{69} & \underline{31.2} & \underline{97.9} & \textbf{35} & \textbf{20.1} & \textbf{3.3} & \underline{243} & 27.4 & \textbf{98.3} & 113 & \underline{29.4} & 94.2 & 93 & 31.2 & \underline{99.2} & 85 \\
\midrule
    \multicolumn{13}{c}{\textit{Erasing \textbf{Snoopy} and \textbf{Mickey}}} & \multicolumn{12}{|c}{\textit{Erasing \textbf{Picasso}}} \\
\midrule
    & \cellcolor{tablehighlight}CS$^{\downarrow}$ & \cellcolor{tablehighlight}CA$^{\downarrow}$ & \cellcolor{tablehighlight}FID$^{\uparrow}$ & \cellcolor{tablehighlight}CS$^{\downarrow}$ & \cellcolor{tablehighlight}CA$^{\downarrow}$ & \cellcolor{tablehighlight}FID$^{\uparrow}$ & CS$^{\uparrow}$ & CA$^{\uparrow}$ & FID$^{\downarrow}$ & CS$^{\uparrow}$ & CA$^{\uparrow}$ & FID$^{\downarrow}$ & CS$^{\uparrow}$ & CA$^{\uparrow}$ & FID$^{\downarrow}$ & \cellcolor{tablehighlight}CS$^{\downarrow}$ & \cellcolor{tablehighlight}CA$^{\downarrow}$ & \cellcolor{tablehighlight}FID$^{\uparrow}$ & CS$^{\uparrow}$ & CA$^{\uparrow}$ & FID$^{\downarrow}$ & CS$^{\uparrow}$ & CA$^{\uparrow}$ & FID$^{\downarrow}$ \\
    \midrule
    ESD-x & \cellcolor{tablehighlight}24.6 & \cellcolor{tablehighlight}\textbf{4.2} & \cellcolor{tablehighlight}\underline{300} & \cellcolor{tablehighlight}25.0 & \cellcolor{tablehighlight}\textbf{30.0} & \cellcolor{tablehighlight}\textbf{305} & 26.7 & 58.3 & 300 & 27.8 & 65.0 & 243 & 24.1 & 78.3 & 265 & \cellcolor{tablehighlight}24.6 & \cellcolor{tablehighlight}78.3 & \cellcolor{tablehighlight}\textbf{282} & 21.9 & 15.4 & 265 & 25.1 & \underline{82.5} & 242 \\
    ESD-u & \cellcolor{tablehighlight}31.6 & \cellcolor{tablehighlight}99.2 & \cellcolor{tablehighlight}58 & \cellcolor{tablehighlight}31.6 & \cellcolor{tablehighlight}97.9 & \cellcolor{tablehighlight}77 & \textbf{30.4} & 90.4 & \underline{85} & \textbf{31.2} & \underline{98.3} & \underline{43} & \underline{28.1} & \underline{96.7} & \underline{45} & \cellcolor{tablehighlight}28.7 & \cellcolor{tablehighlight}97.9 & \cellcolor{tablehighlight}51 & \underline{29.5} & \underline{96.2} & \underline{29} & 31.6 & \textbf{100.0} & \underline{32} \\
    UCE & \cellcolor{tablehighlight}\underline{22.8} & \cellcolor{tablehighlight}\underline{5.0} & \cellcolor{tablehighlight}\textbf{319} & \cellcolor{tablehighlight}27.4 & \cellcolor{tablehighlight}71.2 & \cellcolor{tablehighlight}201 & 29.0 & 83.8 & 138 & 30.0 & 93.8 & 122 & \textbf{28.9} & \textbf{97.1} & \textbf{41} & \cellcolor{tablehighlight}26.7 & \cellcolor{tablehighlight}90.0 & \cellcolor{tablehighlight}92 & \textbf{30.0} & \underline{96.2} & \textbf{28} & \textbf{31.8} & \textbf{100.0} & \textbf{30} \\
    CASteer & \cellcolor{tablehighlight}23.1 & \cellcolor{tablehighlight}45.8 & \cellcolor{tablehighlight}228 & \cellcolor{tablehighlight}\textbf{23.4} & \cellcolor{tablehighlight}56.7 & \cellcolor{tablehighlight}290 & \underline{29.8} & \underline{91.7} & 130 & \textbf{31.2} & \textbf{99.2} & 64 & 24.2 & 95.4 & 185 & \cellcolor{tablehighlight}\underline{21.6} & \cellcolor{tablehighlight}\underline{54.6} & \cellcolor{tablehighlight}207 & \underline{29.5} & 94.2 & 80 & 31.5 & \textbf{100.0} & 68 \\
    \midrule
\rowcolor{ourshighlight}
    \textbf{\ours} & \textbf{22.3} & 14.2 & 231 & \underline{23.6} & \underline{44.6} & \underline{295} & \textbf{30.4} & \textbf{93.3} & \textbf{78} & \underline{31.1} & \underline{98.3} & \textbf{40} & 26.4 & \underline{96.7} & 110 & \textbf{20.9} & \textbf{24.6} & \underline{259} & \textbf{30.0} & \textbf{97.1} & 68 & \underline{31.7} & \textbf{100.0} & 53 \\
\midrule
    \multicolumn{13}{c}{\textit{Erasing \textbf{Snoopy}, \textbf{Mickey} and \textbf{Spongebob}}} & \multicolumn{12}{|c}{\textit{Erasing \textbf{Rembrandt}}} \\
\midrule
    & \cellcolor{tablehighlight}CS$^{\downarrow}$ & \cellcolor{tablehighlight}CA$^{\downarrow}$ & \cellcolor{tablehighlight}FID$^{\uparrow}$ & \cellcolor{tablehighlight}CS$^{\downarrow}$ & \cellcolor{tablehighlight}CA$^{\downarrow}$ & \cellcolor{tablehighlight}FID$^{\uparrow}$ & \cellcolor{tablehighlight}CS$^{\downarrow}$ & \cellcolor{tablehighlight}CA$^{\downarrow}$ & \cellcolor{tablehighlight}FID$^{\uparrow}$ & CS$^{\uparrow}$ & CA$^{\uparrow}$ & FID$^{\downarrow}$ & CS$^{\uparrow}$ & CA$^{\uparrow}$ & FID$^{\downarrow}$ & CS$^{\uparrow}$ & CA$^{\uparrow}$ & FID$^{\downarrow}$ & \cellcolor{tablehighlight}CS$^{\downarrow}$ & \cellcolor{tablehighlight}CA$^{\downarrow}$ & \cellcolor{tablehighlight}FID$^{\uparrow}$ & CS$^{\uparrow}$ & CA$^{\uparrow}$ & FID$^{\downarrow}$ \\
    \midrule
    ESD-x & \cellcolor{tablehighlight}24.7 & \cellcolor{tablehighlight}64.6 & \cellcolor{tablehighlight}232 & \cellcolor{tablehighlight}\underline{26.2} & \cellcolor{tablehighlight}79.2 & \cellcolor{tablehighlight}255 & \cellcolor{tablehighlight}25.9 & \cellcolor{tablehighlight}87.9 & \cellcolor{tablehighlight}262 & 30.5 & 97.1 & 161 & 24.5 & \textbf{99.6} & 525 & 21.8 & 0.4 & 459 & \cellcolor{tablehighlight}20.6 & \cellcolor{tablehighlight}\textbf{0.0} & \cellcolor{tablehighlight}\textbf{520} & 24.9 & \underline{22.9} & 418 \\
    ESD-u & \cellcolor{tablehighlight}32.0 & \cellcolor{tablehighlight}100.0 & \cellcolor{tablehighlight}49 & \cellcolor{tablehighlight}31.6 & \cellcolor{tablehighlight}96.7 & \cellcolor{tablehighlight}70 & \cellcolor{tablehighlight}30.3 & \cellcolor{tablehighlight}92.1 & \cellcolor{tablehighlight}75 & \textbf{31.2} & \underline{97.9} & \textbf{33} & \underline{28.1} & \underline{97.1} & \underline{48} & \underline{28.7} & 97.9 & \underline{47} & \cellcolor{tablehighlight}29.0 & \cellcolor{tablehighlight}90.8 & \cellcolor{tablehighlight}41 & \underline{31.6} & \textbf{100.0} & \underline{37} \\
    UCE & \cellcolor{tablehighlight}23.2 & \cellcolor{tablehighlight}\textbf{9.2} & \cellcolor{tablehighlight}\textbf{304} & \cellcolor{tablehighlight}29.5 & \cellcolor{tablehighlight}83.8 & \cellcolor{tablehighlight}146 & \cellcolor{tablehighlight}24.7 & \cellcolor{tablehighlight}70.0 & \cellcolor{tablehighlight}\textbf{361} & 29.9 & 95.4 & 131 & \textbf{29.0} & \underline{97.1} & \textbf{42} & \textbf{29.4} & \textbf{98.8} & \textbf{40} & \cellcolor{tablehighlight}26.9 & \cellcolor{tablehighlight}78.8 & \cellcolor{tablehighlight}80 & \textbf{31.7} & \textbf{100.0} & \textbf{31} \\
    CASteer & \cellcolor{tablehighlight}\underline{23.1} & \cellcolor{tablehighlight}46.7 & \cellcolor{tablehighlight}229 & \cellcolor{tablehighlight}\textbf{23.4} & \cellcolor{tablehighlight}\underline{54.2} & \cellcolor{tablehighlight}\underline{292} & \cellcolor{tablehighlight}\underline{23.4} & \cellcolor{tablehighlight}\underline{40.0} & \cellcolor{tablehighlight}337 & 30.7 & \underline{97.9} & 83 & 25.1 & 93.3 & 183 & 26.7 & 96.7 & 144 & \cellcolor{tablehighlight}\underline{20.3} & \cellcolor{tablehighlight}26.7 & \cellcolor{tablehighlight}272 & 31.4 & \textbf{100.0} & 66 \\
    \midrule
\rowcolor{ourshighlight}
    \textbf{\ours} & \textbf{22.2} & \underline{19.2} & \underline{240} & \textbf{23.4} & \textbf{47.9} & \textbf{299} & \textbf{22.1} & \textbf{15.8} & \underline{348} & \underline{31.0} & \textbf{98.3} & \underline{55} & 27.4 & 96.2 & 119 & 28.4 & \underline{98.3} & 124 & \textbf{18.4} & \underline{2.5} & \underline{318} & \underline{31.6} & \textbf{100.0} & 60 \\
    \bottomrule
\end{tabular}
\end{table}

We repeat the full six-setting instance and style protocol on Stable Diffusion v1.5, Stable Diffusion
v2.1, and SDXL. The steering vectors and all method-specific weights are re-estimated on each backbone
rather than transferred from SD-v1.4. Tables~\ref{tab:xmodel_sd15}--\ref{tab:xmodel_sdxl} report the
results for every method supported by the corresponding architecture.

The erase--preserve pattern from the main results transfers across all three backbones. Relative to
CASteer, \ours\ generally lowers target accuracy while maintaining higher retained accuracy and lower
retained FID on both instances and styles. The gains are largest on SD-1.x and narrower on SD-v2.1,
where the targets are harder to suppress. On SDXL, the same retain-aware construction transfers to its
dual text encoders and higher-resolution denoiser without changing the projection rule. Methods whose
released implementations do not support SDXL are omitted rather than replaced by custom variants.

\subsection{Additional Qualitative Comparisons}
\label{app:qual}

\begin{figure}[p]
\centering
\includegraphics[width=\linewidth,height=0.86\textheight,keepaspectratio]{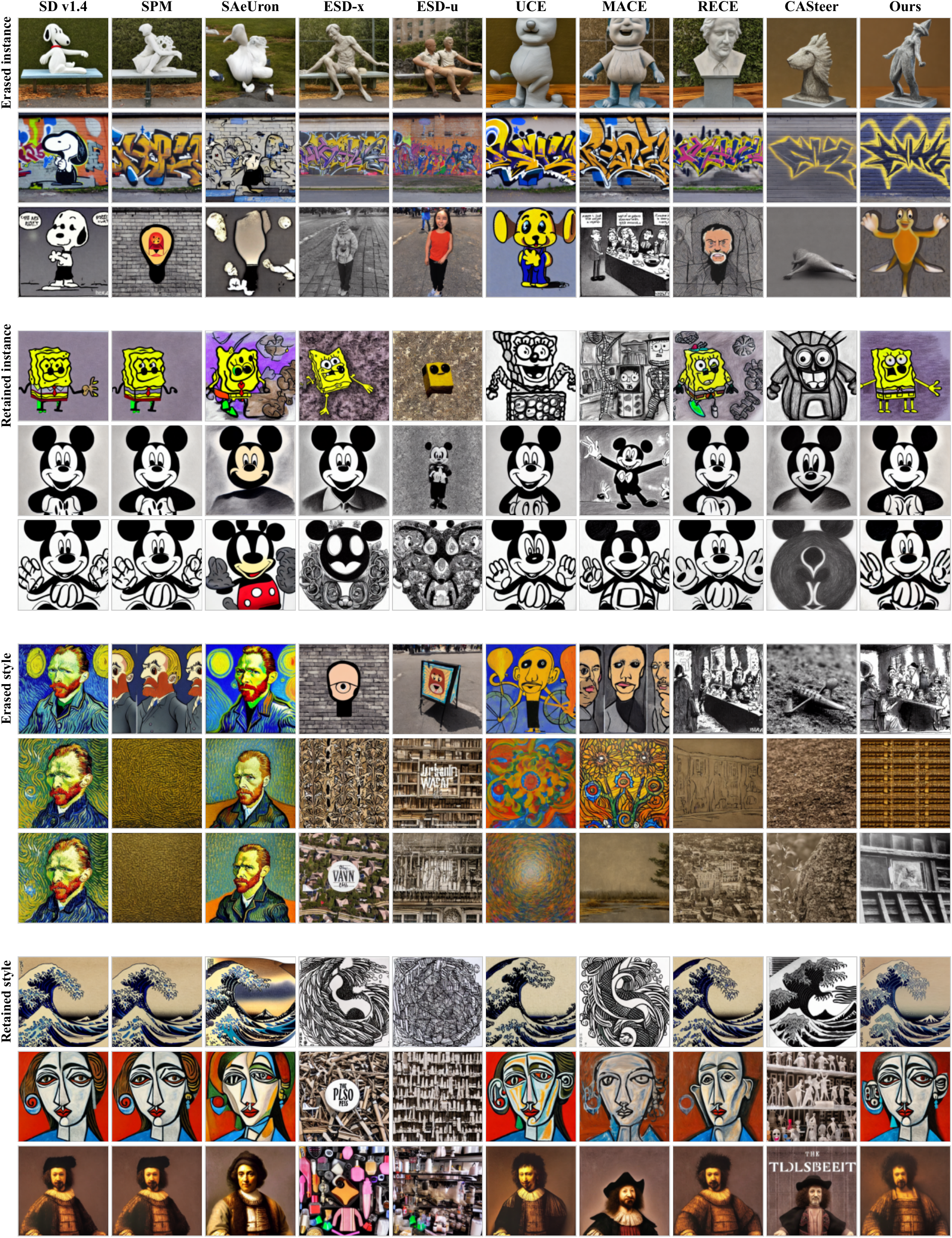}
\caption{Additional instance and style comparisons on Stable Diffusion v1.4. Each group of three rows
shows representative prompts from the indicated erased or retained set.}
\label{fig:qual_sd14}
\end{figure}

\begin{figure}[p]
\centering
\includegraphics[width=\linewidth,height=0.86\textheight,keepaspectratio]{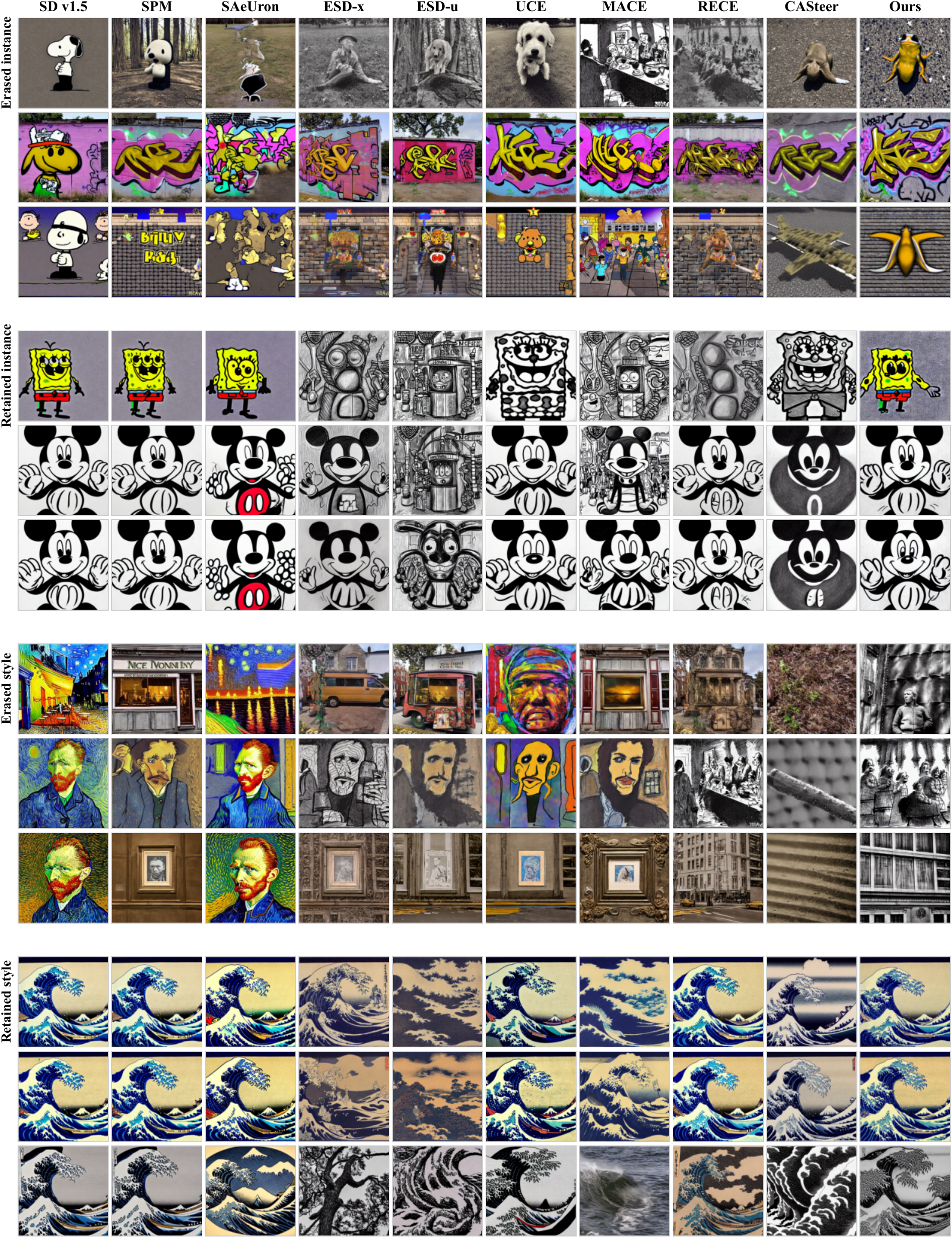}
\caption{Additional instance and style comparisons on Stable Diffusion v1.5. Each group of three rows
shows representative prompts from the indicated erased or retained set.}
\label{fig:qual_sd15}
\end{figure}

\begin{figure}[p]
\centering
\includegraphics[width=\linewidth,height=0.86\textheight,keepaspectratio]{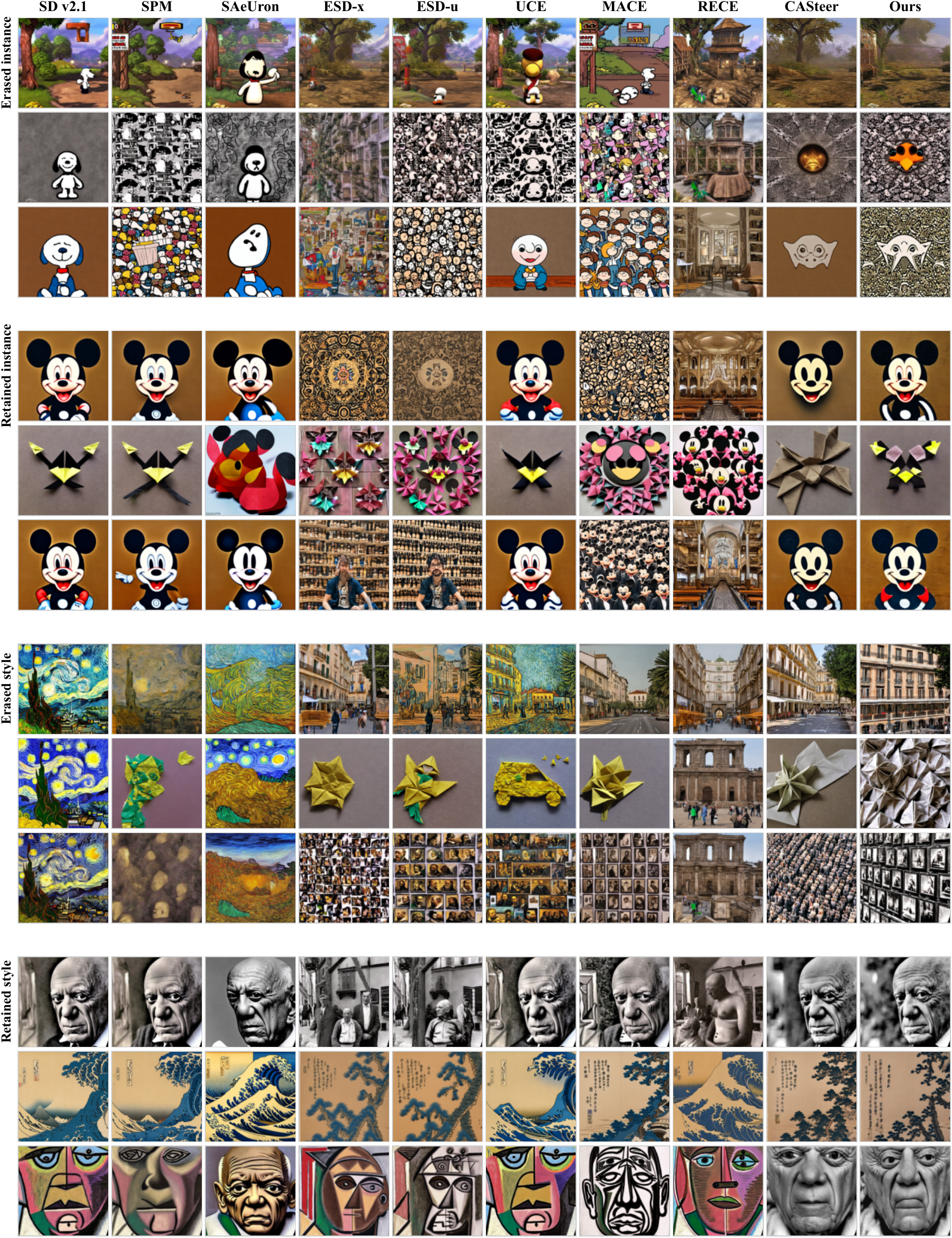}
\caption{Additional instance and style comparisons on Stable Diffusion v2.1. Each group of three rows
shows representative prompts from the indicated erased or retained set.}
\label{fig:qual_sd21}
\end{figure}

\begin{figure}[p]
\centering
\includegraphics[width=\linewidth,height=0.86\textheight,keepaspectratio]{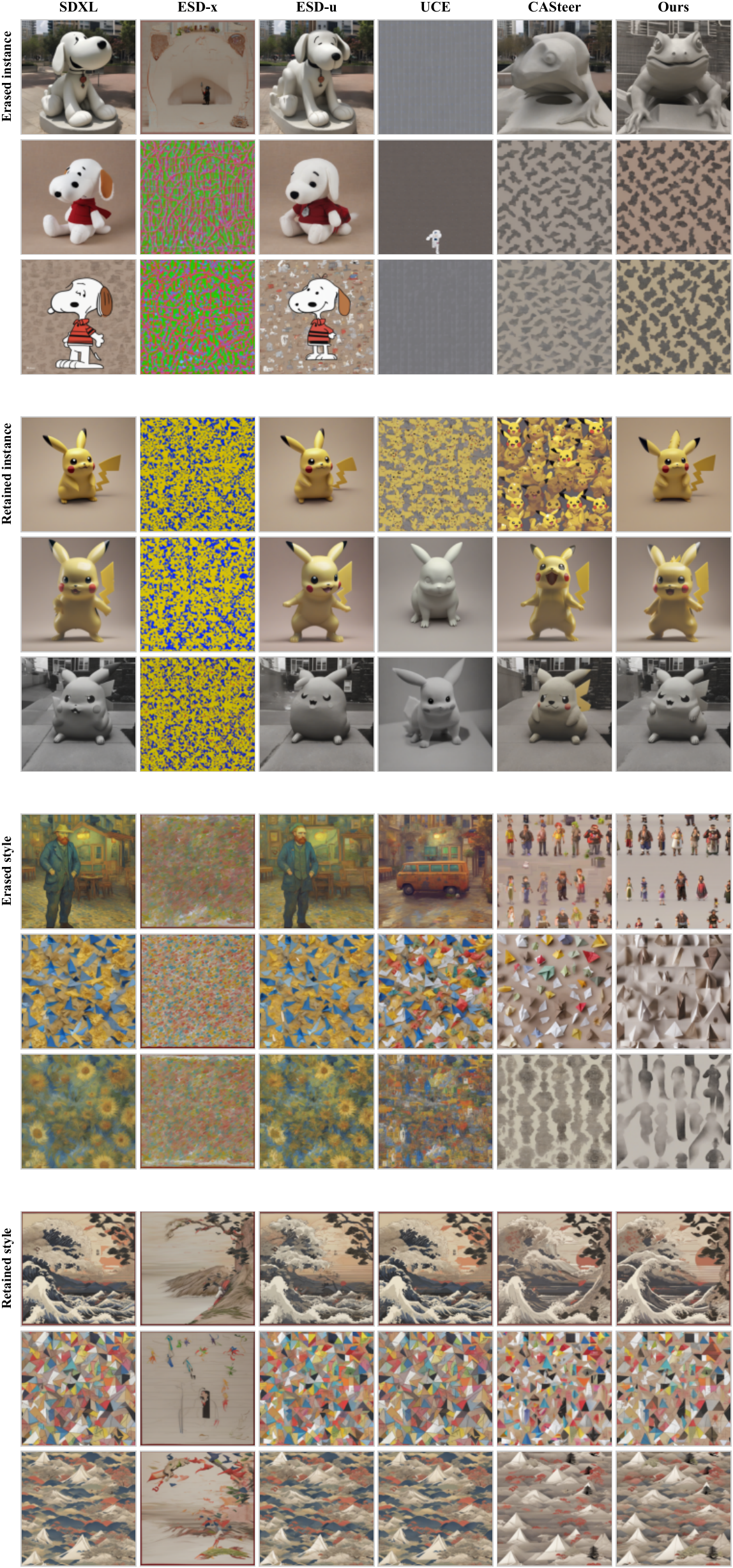}
\caption{Additional instance and style comparisons on SDXL. Each group of three rows shows
representative prompts from the indicated erased or retained set.}
\label{fig:qual_sdxl}
\end{figure}

Figures~\ref{fig:qual_sd14}--\ref{fig:qual_sdxl} extend the main-text comparison to all four
backbones. Every figure uses the same layout: three erased instances, three retained instances, three
erased styles, and three retained styles. Columns show the unedited backbone, each supported baseline,
and \ours. Within a row, all methods use the same prompt and seed; differences can therefore be read
as method effects rather than prompt or sampling variation.

We analyze each row from left to right along two axes. For an erased row, success requires both removal
of the target identity or style and a visually coherent replacement; a recognizable Snoopy or Van Gogh
signature indicates under-erasure, whereas repeated textures, collapsed structure, or an empty scene
indicate over-erasure. For a retained row, we compare identity, composition, color, and texture against
the unedited output in the first column. Large deviations reveal collateral damage even when the erased
target itself is absent.

Across SD-v1.4 and SD-v1.5, \ours\ generally removes the target while producing structured alternatives,
whereas several baselines either retain recognizable target cues or alter nearby concepts. SD-v2.1 is
the more difficult regime: residual target cues occur more often, but the retained Mickey Mouse and
Hokusai rows remain comparatively stable under \ours. On SDXL, target removal is strong and the retained
Pikachu and artistic-style rows expose the preservation advantage most clearly; the remaining texture
changes in erased-style rows also show that strong erasure is not artifact-free. These recurring patterns
match the quantitative erasure--preservation trade-off rather than relying on a single favorable sample.

A backbone-specific reading further separates the failure modes. On SD-v1.4 and SD-v1.5, incomplete
instance erasure often leaves a recognizable silhouette, while stronger interventions can replace the
entire subject or background. In the style rows, some methods suppress the named painter by flattening
the image into repetitive texture rather than preserving the prompt content. SD-v2.1 shows the same
effects more strongly: residual character shapes and dense repeated motifs coexist across methods. The
SDXL rows make collateral damage especially easy to see because its unedited outputs are highly
consistent; the retained Pikachu poses and wave-like compositions stay close under \ours, whereas some
baselines alter the subject, palette, or spatial structure. Thus the grid distinguishes successful
semantic removal from mere visual degradation.

\clearpage

\clearpage

\clearpage

\clearpage
\clearpage

\end{document}